\documentclass{article}

\usepackage[verbose=true,letterpaper]{geometry}
  \newgeometry{
    textheight=9in,
    textwidth=5.5in,
    top=1in,
    headheight=12pt,
    headsep=25pt,
    footskip=30pt
  }

\usepackage{parskip}

\usepackage{natbib}
\PassOptionsToPackage{dvipsnames}{xcolor}

\usepackage{hyperref}

\usepackage[utf8]{inputenc} 
\usepackage[T1]{fontenc}    
\usepackage{url}            
\usepackage{booktabs}       
\usepackage{amsfonts}       
\usepackage{nicefrac}       
\usepackage{microtype}      
\usepackage{subcaption}
\usepackage{amsmath}
\usepackage{tabularx}
\usepackage{mathabx}
\usepackage{amssymb}
\usepackage{enumitem}
\usepackage{cleveref}
\usepackage{wrapfig}
\usepackage{multirow}
\usepackage{tcolorbox}
\usepackage{adjustbox}
\usepackage{makecell}

\usepackage{hyperref}       
\usepackage{url}            
\usepackage{booktabs}       
\usepackage{amsfonts}       
\usepackage{nicefrac}       
\usepackage{microtype}      
\usepackage{xcolor}         
\usepackage{color, colortbl}
\usepackage{array}
\usepackage{graphicx}
\usepackage{multirow}
\usepackage{amsmath}
\usepackage{amssymb}
\usepackage{pifont}
\usepackage{xspace}
\usepackage{wrapfig}
\usepackage{tabularx}
\usepackage{float}
\usepackage{longtable}
\usepackage{adjustbox}
\usepackage{caption}
\usepackage{listings}
\usepackage{comment}
\definecolor{vscText}{rgb}{0.0, 0.0, 0.0}          
\definecolor{vscKeyword}{rgb}{0.0, 0.0, 0.6}       
\definecolor{vscString}{rgb}{0.0, 0.5, 0.0}        
\definecolor{vscComment}{rgb}{0.5, 0.5, 0.5}       
\definecolor{vscFunction}{rgb}{0.6, 0.0, 0.6}      
\definecolor{vscNumber}{rgb}{0.6, 0.2, 0.2}        
\definecolor{LightCyan}{rgb}{0.88,1,1}
\lstdefinelanguage{Python}{
    keywords={def, return, if, elif, else, for, while, break, continue, pass, import, from, as, class, try, except, finally, raise, with, lambda},
    keywordstyle=\color{vscKeyword}\bfseries,
    ndkeywords={self},
    ndkeywordstyle=\color{vscKeyword}\bfseries,
    identifierstyle=\color{vscFunction},
    sensitive=false,
    comment=[l]{\#},
    morecomment=[s]{"""}{"""},
    commentstyle=\color{vscComment}\ttfamily,
    stringstyle=\color{vscString}\ttfamily,
    morestring=[b]',
    morestring=[b]",
    morestring=[s]{"""}{"""}
}

\newcommand{\xmark}{\ding{55}}%

\newcommand{\vs}[1]{\textcolor{blue}{\bf\small [Vedaant]}}

\hypersetup{
    colorlinks=true,
    linkcolor=blue,
    filecolor=magenta,      
    urlcolor=magenta,
    citecolor=blue,
    pdftitle={Translation and Fusion Improves Zero-shot Cross-lingual Information Extraction},
    pdfpagemode=FullScreen,
    }

\definecolor{lightblue}{HTML}{84C7F9}
\definecolor{lighterblue}{HTML}{D4ECFF}
\newtcolorbox{mybox}{colback=lighterblue,colframe=lightblue}

\usepackage[toc,page,header]{appendix}
\usepackage{minitoc}

\date{}

\title{Translation and Fusion Improves \\ Zero-shot Cross-lingual Information Extraction}

\author{
\normalsize \textbf{Yang Chen}\thanks{ Correspondence to: ychen3411@gatech.edu} 
\quad \textbf{Vedaant Shah} \quad \textbf{Alan Ritter} \\
\normalsize Georgia Institute of Technology
}

\begin{document}

\doparttoc
\faketableofcontents 

\maketitle

\begin{abstract}
\noindent
Large language models (LLMs) combined with instruction tuning have shown significant progress in information extraction (IE) tasks, exhibiting strong generalization capabilities to unseen datasets by following annotation guidelines.
However, their applicability to low-resource languages remains limited due to lack of both labeled data for fine-tuning, and unlabeled text for pre-training.
In this paper, we propose TransFusion, a framework in which models are fine-tuned to use English translations of low-resource language data, enabling more precise predictions through annotation fusion. 
Based on TransFusion, we introduce GoLLIE-TF, a cross-lingual instruction-tuned LLM for IE tasks, designed to close the performance gap between high and low-resource languages.
Our experiments across twelve multilingual IE datasets spanning 50 languages demonstrate that GoLLIE-TF achieves better zero-shot cross-lingual transfer over the base model.
In addition, we show that TransFusion significantly improves low-resource language named entity recognition when applied to proprietary models such as GPT-4 (+5 F1) with a prompting approach, or fine-tuning different language models including decoder-only (+14 F1) and encoder-only (+13 F1) architectures.\footnote{Code and data is available at: \url{https://github.com/edchengg/gollie-transfusion}} 
\end{abstract}

\section{Introduction}

\begin{figure*}[t!]
\centering
\includegraphics[width=\textwidth]{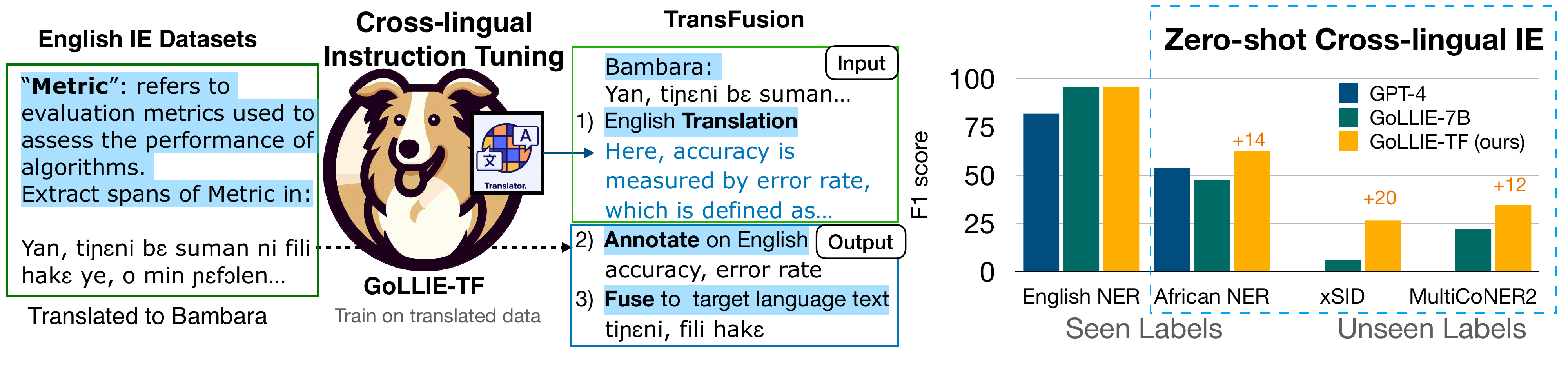}
\caption{Our TransFusion framework aims to bridge the performance gap between high and low-resource languages on information extraction tasks. (left) TransFusion reasoning includes three steps: translate, annotate, and fuse. (right) GoLLIE-TF shows superior zero-shot cross-lingual evaluation on a range of IE datasets over the base model.}
\label{fig:gollie-tf}
\end{figure*}

The task of information extraction (IE) is challenging due to fine-grained annotation guidelines for span-level annotations. 
Fortunately, recent advances in instruction-following large language models (LLM)~\citep{instructgpt,team2023gemini} such as GoLLIE~\citep{sainz2024gollie} have demonstrated the ability to perform zero-shot IE without labels using annotation guidelines.
However, these models are often pre-trained on English-centric data \citep{touvron2023llama,roziere2023codellama}.  Even state-of-the-art proprietary models such as GPT-4 exhibit significant performance degradation from 80 English F1 to 55 F1 on low-resource African languages, as shown in Figure~\ref{fig:gollie-tf} (right).

To improve NLP on low-resource languages, the research community has turned to machine translation to translate fine-tuning datasets (translate-train) and translate test data into high-resource languages for easier processing (translate-test)~\citep{hu2020xtreme}. Recent studies~\citep{shi2022language,huang2023not} on prompting LLMs with translated data have shown improvements on diverse tasks such as math reasoning and summarization. 
Prior work has explored the use of machine translation to improve multilingual instruction-following on traditional NLP benchmarks, such as natural language inference, and sentiment analysis, however, the use of MT to improve instruction-following IE models is less explored, as there is not a trivial alignment between labels in the native language and translated texts~\citep{ahuja2023mega}. 
With recent efforts to develop machine translation (MT) models such as M2M~\citep{fan2021beyond} and NLLB-200~\citep{nllb2022} that better support low-resource languages, we study how to teach LLMs to leverage an external MT system in a resource-efficient manner to improve low-resource IE.

In this paper, we propose a Translation and Fusion (TransFusion) framework, which aims to teach models to use an external translation model to make better predictions on low-resource languages.
The framework includes three steps: (1) translating low-resource data into English at inference time, to be annotated by a high-resource model.  Next, (2) these span-annotated English translations are combined with low-resource language text in a fusion model that is trained to make predictions conditioned on both types of data.  Finally (3), the language model generates a TransFusion reasoning chain (annotate and fuse) in a single autoregressive decoding pass.
To train TransFusion models, we construct cross-lingual instruction fine-tuning data by translating and projecting labels from English IE datasets to low-resource languages using EasyProject~\citep{chen2022frustratingly}, a simple, yet effective method that has been shown to scale across many NLP tasks and languages.

Our zero-shot cross-lingual IE evaluation reveals that the TransFusion fine-tuned model, GoLLIE-TF, outperforms the base GoLLIE model across 50 languages, spanning high, mid, and low-resource categories, on both seen and unseen label schemas. Notably, in our evaluation on African language named entity recognition (NER) using the MasakhaNER2 dataset~\citep{adelani2022masakhaner}, GoLLIE-TF achieves significant improvements in F$_1$ scores compared to the base model and shows an average improvement of +6.6 F$_1$ on unseen label schema datasets. 
Furthermore, we demonstrate that the TransFusion framework enhances GPT-4's performance on MasakhaNER2, yielding an average +5.7 F$_1$ score improvement, and substantially boosts the encoder-only African language model, AfroXLM-R~\citep{alabi-etal-2022-adapting}, by +13.3 F$_1$.  Our analysis underscores the effectiveness of the TransFusion framework for low-resource language tasks.

\section{Background: Annotation Guideline Following LLMs for IE}
In this paper, we employ the GoLLIE model~\citep{sainz2024gollie}, which has been instruction-tuned on English Information Extraction (IE) tasks using label schema guidelines, to achieve state-of-the-art zero-shot IE on unseen datasets. GoLLIE utilizes a Python code representation for both inputs and outputs, providing a clear and human-readable structure that unifies various IE annotation tasks. Each label schema is encapsulated as a Python class object, with the annotation guidelines embedded as strings within these objects (an example of a GoLLIE prompt is provided in the the Appendix in Figure~\ref{fig:x-gollie}.

\textbf{Limitation of Cross-lingual 
Transferbilitiy}: Despite GoLLIE's impressive performance, it is designed for use on English, as it is primarily fine-tuned on English data. This limitation is shown in Figure~\ref{fig:gollie-tf} (right), where we see a significant drop in performance on low-resource African languages, from 95 to 48, compared to English.
In this study, we experiment with \textbf{zero-shot cross-lingual transfer}, where human-labeled data in the target languages are assumed to be unavailable. Collecting such data is costly and time-inefficient, as it requires well-trained native language speakers. 
Recent efforts, such as NLLB-200~\citep{nllb2022}, have focused on gathering low-resource translation data to train multilingual MT models capable of translating across 200 languages.  Building on this progress, we explore whether an instruction-tuned information extraction model can learn to use an external translation model~\citep{schick2024toolformer} to enhance performance on low-resource language IE tasks. This offers an efficient and effective alternative to computationally intensive pre-training based methods for adapting to new languages~\citep{scao2022bloom, xue2021mt5, alabi-etal-2022-adapting, ustun2024aya}.

\section{Using Low-Resource Machine Translation to Improve Multilingual IE}

As multilingual machine translation (MT) systems, such as M2M-100~\citep{fan2021beyond} and NLLB-200~\citep{nllb2022}, gain increasing support for low-resource languages, an opportunity emerges to re-evaluate the utilization of MT systems for enhancing cross-lingual IE.
We propose a Translation-and-fusion approach that benefits from the advancements of MT systems to make robust cross-lingual transfer predictions at inference time.
In this section, we outline the Translation-and-fusion framework and introduce language models trained to utilize translation data at inference time for low-resource language IE tasks.

\subsection{Translation-and-Fusion (TransFusion)}
\label{sec:fusion}
\paragraph{Cross-lingual Transfer.} The conventional cross-lingual transfer method involves fine-tuning a pre-trained language model, on high-resource language annotated data {\small ($src$)} and evaluating its performance on test data in other languages {\small $(tgt)$}.  

In accordance with the low-resource assumption, we assume access to an annotated dataset in the high-resource language (usually English), $\mathcal{D}_{src} = {(x_{src}^i, y_{src}^i)}_{i=1}^N$. 
The task-specific fine-tuning loss is formulated as:
\begin{equation*}
\mathcal{L}(\theta, \mathcal{D}_{src}) = 
\sum_{(x_{src}, y_{src}) \in \mathcal{D}_{src}} \mathcal{L}(P(y|x_{src}; \theta), y_{src})
\end{equation*}

However, previous studies have highlighted the limited performance of fine-tuned models on languages that were unseen during pre-training or are under-represented in the pre-training data \citep{adelani2021masakhaner, ebrahimi2021americasnli}. As an additional approach to adapt to low-resource languages~\citep{wang-etal-2020-extending}, we describe the translation-and-fusion framework, which leverages annotations on (translated) high-resource language text to steer predictions on a low-resource language at inference time. The framework encompasses three key steps:
\begin{itemize}
\item \textbf{Translate}: Use an MT system to translate low-resource language test data into a high-resource language, $\text{MT}(x_{tgt}) \mapsto x^{\text{\scriptsize{trans}}}_{src}$.
\item \textbf{Annotate}: Making predictions to the (high-resource) translated text using a high-resource language supervised fine-tuned model $P(;\theta_{src})$:  
$\mathrm{argmax}_y\{P(y|x^{\text{\scriptsize{trans}}}_{src}; \theta_{src})\} \mapsto \Tilde{y}^{\text{\scriptsize{trans}}}_{src}$.
\item \textbf{Fuse}:

Given the annotations on the translated data from the previous step ($\Tilde{y}^{\text{\scriptsize{trans}}}_{src}$), a fusion model combines the \textit{high-resource predictions} together with the target language text to make final predictions.
\end{itemize}

Based on the framework outlined above, we present TransFusion, a fusion model that is trained to makes predictions on the test data conditioned on annotations from the corresponding translated data ($\Tilde{y}^{\text{\scriptsize{trans}}}_{src}$):
\begin{equation*}
\mathrm{argmax}_y\{P(y|x_{tgt}, x^{\text{\scriptsize{trans}}}_{src}, \Tilde{y}^{\text{\scriptsize{trans}}}_{src}; \theta_\text{fusion})\} \mapsto y_{tgt}'
\end{equation*}
\noindent

Below, we describe the training procedure of TransFusion, starting with the approach to create training data.

\paragraph{Training Dataset.}
To learn a TransFusion model, parallel sentences with IE task annotations on both high-resource and low-resource languages are essential.
To fulfill this requirement, we translate high-resource annotated training data into a list of target languages, while projecting span-level annotations, using a simple mark-then-translate approach - EasyProject~\citep{chen2022frustratingly}: MT$(x_{src}, y_{src}) \to (x^{\text{\scriptsize{trans}}}_{tgt}, y^{\text{\scriptsize{trans}}}_{tgt})$. 
We then pair the translation outputs with the original high-resource language data to create a training data set with a mixture of both parallel sentences: $\mathcal{D}_{mix} = \{x_{src}, y_{src}, x^{\text{\scriptsize{trans}}}_{tgt}, y^{\text{\scriptsize{trans}}}_{tgt}\}^N_{i=1}$.

\paragraph{Learning.}
We train the fusion model $P(;\theta_\text{fusion})$ on $\mathcal{D}_{mix}$ using cross-entropy loss:
\begin{equation*}
\mathcal{L}_{\text{fusion}}(\theta, \mathcal{D}_{mix}) = 
\sum_{\substack{
  (x_{src}, y_{src}, x^{\text{\scriptsize{trans}}}_{tgt}, y^{\text{\scriptsize{trans}}}_{tgt})
  \in \mathcal{D}_{mix}
  }}
\mathcal{L}(P(y|x^{\text{\scriptsize{\text{\scriptsize{trans}}}}}_{tgt}, x_{src}, y_{src}; \theta_{\text{fusion}}), y^{\text{\scriptsize{trans}}}_{tgt})
\end{equation*}

The model architecture can vary, encompassing both decoder-only language models (e.g., LLaMA~\citep{touvron2023llama}) and encoder-only language models (e.g., mBERT~\citep{devlin2019bert}). In this work, we primarily utilize decoder-only language models to integrate the \textit{annotate} and \textit{fuse} steps in an autoregressive manner during inference. Additionally, we assess the performance of encoder-only models in Section~\ref{sec:encoder} to demonstrate the robustness of our framework across different architectures.

\paragraph{Training a Decoder-only LM (GoLLIE-TF).}
To implement our TransFusion framework within the instruction-following GoLLIE model, we represent the framework as natural language instructions, providing the model with supplementary English translation text of the original target language sentence, which is illustrated in Figure~\ref{fig:gollie-tf} (left). The TransFusion instruction specifies the output format, guiding the model to first generate annotations for the English translation and subsequently for the target language data, using the English annotations as context (an example can be found in Appendix Figure~\ref{fig:x-gollie}
). This autoregressive approach enables the model to perform the annotate and fuse steps concurrently during inference.
During training, we fine-tune the GoLLIE model to adhere to these instructions, ensuring it generates annotations for both the English and target language data sequentially. We apply the next token prediction loss to the tokens following the TransFusion instruction, as detailed below:
\begin{equation*}
    [\texttt{GoLLIE Guidelines}, x, x^{trans}, \texttt{TransFusion Instruction}] \stackrel{\text{LLM}}{\xrightarrow{\hspace*{1cm}}} [y^{trans}, y]
\end{equation*}

\paragraph{Training and Inference with Encoder-only LMs.}

Given that encoder-only models are not inherently designed for text generation, we employ a two-step pipeline approach for inference in TransFusion: annotation and fusion.
First, we utilize an English fine-tuned model to annotate the English translation of the target language text. These annotations are marked using XML tags around the relevant spans (e.g., \texttt{<PER>} ... \texttt{</PER>}).
Next, we construct the input for the fusion model by embedding these annotations into the English translation. We concatenate the annotated English translation ($x^{trans}$) with the original target language text ($x$), using a marker ($||$) to separate the two segments. The input to the encoder is formatted as follows:

\begin{equation*}
    [x^{trans}_1, x^{trans}_2, \texttt{<PER>}, x^{trans}_3, x^{trans}_4, \texttt{</PER>}, x^{trans}_5, ||, x_1, x_2, x_3, ...]
\end{equation*}
At training time, we add a linear classification layer to classify each token and only apply the cross-entropy loss to the target language tokens (right of the separation token $||$).

\section{Experimental Setting}
\begin{table*}[ht!]
\centering
\small
\vspace{0.2cm}

\begin{tabularx}{\textwidth}{l|c|c|l}
\toprule
 \textbf{Training Dataset} & Domain & Tasks &  Language\\
\midrule
CoNLL 03\citep{sang2003conll} & News & NER& English \\
BC5CDR~\citep{Li2016BioCreativeVC} & Biomedical & NER & English  \\
NCBIDisease~\citep{Dogan2014NCBIDC}  & Biomedical & NER& English  \\
OntoNotes 5~\citep{pradhan-etal-2013-towards}  & News & NER& English  \\
WNUT 2017~\citep{derczynski-etal-2017-results}  & News & NER& English  \\
RAMS~\citep{ebner-etal-2020-multi} & News & Arg. Extraction& English  \\
TACRED~\citep{zhang-etal-2017-position}  & News & Slot Filling & English  \\
CoNLL 04~\citep{roth-yih-2004-linear} & News & Relation Extraction & English  \\
ACE~\citep{walker2006ace} & News & EE, EAE, NER, RE & English \\
\bottomrule
\end{tabularx}

\vspace{0.2cm}

\begin{tabularx}{\textwidth}{l|c|c|c|l}
\midrule
\textbf{Evaluation Dataset} & Domain & Tasks & Seen & \# Language\\
&&&Label?&\\
\midrule
MasakhaNER2.0~\citep{adelani2022masakhaner} & News & NER & \checkmark & 20 African langs\\
UNER~\citep{mayhew2023uner} & News & NER & \checkmark & 13 langs \\
ACE~\citep{walker2006ace} & News & EE, EAE, NER, RE & \checkmark & 3 (en, ar, zh) \\
\midrule
MultiNERD~\citep{tedeschi-navigli-2022-multinerd} & Wikipedia & NER & \xmark & 10 langs\\
MultiCoNER2~\citep{fetahu2023multiconer} & Wikipedia & NER & \xmark & 12 langs\\
xSID~\citep{van-der-goot-etal-2021-xsid} & Dialog &  Slot Detection & \xmark & 10 langs\\
MultiTO~\citep{Schuster2018CrosslingualTL} & Dialog &  Slot Detection & \xmark & 3 (en, es, th)\\ 
Massive~\citep{fitzgerald2022massive} & Dialog &  Slot Detection & \xmark & 15 low-res langs\\ 
RED-FM~\citep{cabot2023redfm} & Wikipedia & Relation Extraction & \xmark & 7 langs\\

\bottomrule
\end{tabularx}
\caption{Datasets used in the experiment. The table shows the task, domain, whether it was used in the training and evaluation including the number of languages in the evaluation set.
}
\label{table:dataset}
\end{table*}
We use a collection of English Information Extraction (IE) datasets for supervised fine-tuning and multilingual IE datasets for evaluation (see Table~\ref{table:dataset}).
Assessing cross-lingual transfer capabilities requires IE datasets annotated in a diverse set of languages. To this end, we gather multilingual Named Entity Recognition (NER) datasets from MasakhaNER2.0~\citep{adelani2022masakhaner} (20 African languages) and UNER~\citep{mayhew2023uner} (13 languages) to conduct low-resource language evaluation on label schemas that are seen during fine-tuning. In addition, we evaluate on unseen label schemas using the non-English subset of ACE2005~\citep{sang2003conll} (Chinese and Arabic), which includes several tasks: NER, RE, Event Extraction (EE), and Event Argument Extraction (EAE).
For evaluation on labels that were unseen during fine-tuning, we use MultiNERD~\citep{tedeschi-navigli-2022-multinerd} (10 high-resource languages), MultiCoNER2 (12 high-resource languages)~\citep{fetahu2023multiconer}, in addition to Slot Intent Detection data from MultiTO~\citep{Schuster2018CrosslingualTL}, xSID (10 high-resource languages)~\citep{van-der-goot-etal-2021-xsid}, 
a subset of Massive (15 low-resource languages were determined based on the NLLB categorization~\citep{nllb2022})~\citep{fitzgerald2022massive} and Relation Extraction (RE) data from RED-FM (7 high-resource languages)~\citep{cabot2023redfm}.
We adopt the data pre-processing and task formulation methodologies used by GoLLIE and use publicly available English training data from GoLLIE to train the model. Due to the high cost of inference with GPT-4, we use 200 examples per language, per task, for evaluation.

\textbf{Multilingual Translation Data.}
The TransFusion framework relies on a machine translation system as a core component. In this paper, we utilize the state-of-the-art open-source multilingual translation model - NLLB-200~\citep{nllb2022}, which has 3.3 billion parameters and supports translation between 200 languages. The NLLB-200-3.3B model translates target language test data into English at test time. For TransFusion training data, a marker-based translation approach named EasyProject~\citep{chen2022frustratingly}, powered by the NLLB-200 model, translates English training data into a collection of 36 target language candidates. 
From this translated data, 8 examples per language and each task are randomly sampled, resulting in around 20-40 examples per language. 
In total, 20,000 examples are used, combining English (19,109) and translated data (891), to train the TransFusion model (See per task and per language distribution in Appendix Figure~\ref{fig:transfusion_stats}). 
This small portion of translation data ensures that the GoLLIE model generalizes to unseen labels while maintaining English performance.

\subsection{Language Models and Baselines}
\paragraph{Models:}
We adopt GoLLIE-7B as our primary starting checkpoint. GoLLIE is an instruction fine-tuned version of CodeLLaMA~\citep{roziere2023codellama} that is trained on approximately 500,000 English demonstrations. Although the model was not explicitly pre-trained on multilingual data, its pre-training corpus includes a substantial amount of high-resource language content, such as Wikipedia, covering a diverse linguistic range~\citep{touvron2023llama}. This makes GoLLIE-7B an appropriate testbed for examining the adaptation of English-centric LLMs to low-resource languages that may be underrepresented in pre-training.
In addition to this decoder-only LLM, we explore encoder-only models specifically pre-trained on African languages, such as AfroXLM-R~\citep{alabi-etal-2022-adapting} in Section~\ref{sec:encoder}. 

\paragraph{Training Setup:}
Initilized from GoLLIE-7B, we continue fine-tuning the model on a dataset of 20,000 TransFusion training examples using QLoRA~\citep{dettmers2024qlora}.  QLoRA has been shown to better maintain the base model's performance~\citep{biderman2024lora} and offers faster training times compared to full fine-tuning.
To implement this, we freeze the transformer model weights and apply LoRA~\citep{hu2021lora} to all linear layers within all the transformer blocks. We set the LoRA rank to 128 and the alpha parameter to 16 based on preliminary experiments as we found smaller alpha leads to more stable training and higher rank for fast convergence.
We use the AdamW optimizer~\citep{kingma2014adam} with a batch size of 16 and a learning rate of 1e-4, managed by a cosine scheduler. The training process was conducted on a setup of 2 NVIDIA A40 GPUs, each equipped with 48GB of memory. The entire experiment session spanned approximately 6 hours. We use greedy decoding at inference time.

\paragraph{Baselines:}
We compare to both the base GoLLIE model, in addition to GPT-4, which represents a state-of-the-art proprietary model pre-trained on multilingual corpora~\citep{achiam2023gpt}. 
We report few-shot prompting results using GPT-4 (\texttt{gpt4-02-14}) with a GoLLIE style prompt. 
Additionally, we explore the application of the TransFusion framework to GPT-4 in Section~\ref{sec:gpt4}.
Furthermore, we use Translate-train (\textbf{Trans-train})~\citep{hu2020xtreme} as another baseline, which shows strong improvements over English fine-tuned (English FT) models~\citep{chen2022frustratingly}. We use the same translated training data used by TransFusion and fine-tune GoLLIE-7B on a total of 20,000 examples (English + translated data).

\begin{table}[t!]
\centering
\small
\vspace{2mm}
\begin{tabularx}{\textwidth}{llcccl}
\toprule
\textbf{Task} & \textbf{Benchmark} & \textbf{GPT-4} & \textbf{GoLLIE}$_{\text{7B}}$ & \textbf{Trans-Train} & \textbf{GoLLIE-TF}\\

\midrule
\multicolumn{6}{l}{\textbf{Seen Label Schema}}\\
\midrule
NER & MasakhaNER2 {\scriptsize (20 languages)} & \\
\cmidrule(lr){2-6}
& Bambara & 42.2 & 38.9 & 40.1 & \textbf{54.8} ({\scriptsize \textcolor{blue}{+15.9}}) \\ 
& Ghomala & \textbf{58.2} & 43.7 & 49.2 & 50.2 ({\scriptsize \textcolor{blue}{+6.5}}) \\ 
& Ewe & 72.2 & \textbf{74.0} & 73.1 & 73.2 ({\scriptsize \textcolor{black}{-0.8}}) \\ 
& Fon & 39.4 & 49.7 & 55.7 & \textbf{57.9} ({\scriptsize \textcolor{blue}{+8.2}}) \\ 
& Hausa & 65.9 & 57.1 & 55.6 & \textbf{67.1} ({\scriptsize \textcolor{blue}{+10.0}}) \\ 
& Igbo & 42.2 & 51.1 & 42.4 & \textbf{56.6} ({\scriptsize \textcolor{blue}{+5.5}}) \\ 
& Kinyarwanda & 47.5 & 45.0 & 47.7 & \textbf{58.5} ({\scriptsize \textcolor{blue}{+13.6}}) \\ 
& Luganda & 62.5 & 61.8 & 66.8 & \textbf{75.5} ({\scriptsize \textcolor{blue}{+13.7}}) \\ 
& Luo & 47.2 & 36.5 & 42.8 & \textbf{51.7} ({\scriptsize \textcolor{blue}{+15.3}}) \\ 
& Mossi & 43.2 & 45.1 & 46.1 & \textbf{48.8} ({\scriptsize \textcolor{blue}{+3.7}}) \\ 
& Chichewa & 71.1 & 39.1 & 59.8 & \textbf{78.2} ({\scriptsize \textcolor{blue}{+39.1}}) \\ 
& Naija & 78.9 & 75.9 & 74.9 & \textbf{81.1} ({\scriptsize \textcolor{blue}{+5.2}}) \\ 
& Shona & 39.5 & 39.7 & 50.4 & \textbf{57.4} ({\scriptsize \textcolor{blue}{+17.6}}) \\ 
& Swahili & \textbf{79.2} & 66.9 & 68.3 & 73.5 ({\scriptsize \textcolor{blue}{+6.5}}) \\ 
& Tswana & 56.3 & 52.1 & 58.9 & \textbf{71.0} ({\scriptsize \textcolor{blue}{+18.9}}) \\ 
& Twi & 44.2 & 41.7 & 50.6 & \textbf{74.2} ({\scriptsize \textcolor{blue}{+32.5}}) \\ 
& Wolof & 52.6 & 49.1 & 55.5 & \textbf{61.9} ({\scriptsize \textcolor{blue}{+12.8}}) \\ 
& Xhosa & 49.8 & 29.2 & 47.6 & \textbf{49.9} ({\scriptsize \textcolor{blue}{+20.7}}) \\ 
& Yoruba & \textbf{54.7} & 35.7 & 39.3 & 54.4 ({\scriptsize \textcolor{blue}{+18.7}}) \\ 
& Zulu & 36.9 & 25.6 & 31.7 & \textbf{52.8} ({\scriptsize \textcolor{blue}{+27.2}}) \\ 
\cmidrule(lr){2-6}
& Average & 54.2 & 47.9 & 52.8 & \textbf{62.4} ({\scriptsize \textcolor{blue}{+14.5}}) \\ 
\midrule
 NER & UNER {\scriptsize (13 languages)} & 69.0 & 73.6 & 73.6 & \textbf{77.8} ({\scriptsize \textcolor{blue}{+4.2}}) \\ 

 NER & ACE05 {\scriptsize  (English, Arabic, Chinese)} & 41.6 & 58.7 & 61.2 & \textbf{61.5} ({\scriptsize \textcolor{blue}{+2.8}}) \\ 
Arg. Extraction & ACE05 {\scriptsize  (English, Arabic, Chinese)} & 11.7 & 92.7 & \textbf{92.9} & 86.0 ({\scriptsize \textcolor{black}{-6.7}}) \\ 
Event Detection & ACE05 {\scriptsize  (English, Arabic, Chinese)} & 21.3 & 42.6 & 40.0 & \textbf{44.0} ({\scriptsize \textcolor{blue}{+1.4}}) \\ 
Rel. Extraction & ACE05  {\scriptsize  (English, Arabic, Chinese)} & 4.6 & 37.3 & \textbf{39.4} & 39.1 ({\scriptsize \textcolor{blue}{+1.8}}) \\ 
\midrule 
 \multicolumn{6}{l}{\textbf{Unseen Label Schema}} \\ 
 \midrule 
NER & MultiNERD {\scriptsize (10 languages)} & \textbf{71.9} & 62.2 & 63.9 & 63.0 ({\scriptsize \textcolor{blue}{+0.8}}) \\ 
 NER & MultiCoNER2 {\scriptsize (12 languages)} & \textbf{46.1} & 22.2 & 28.4 & 34.5 ({\scriptsize \textcolor{blue}{+12.2}}) \\ 
Slot Detection & xSID {\scriptsize (10 languages)} & \textbf{47.0} & 6.0 & 27.1 & 26.4 ({\scriptsize \textcolor{blue}{+20.4}}) \\ 
 Slot Detection & MultiTO {\scriptsize (English, Spanish, Thai)} & 19.9 & 17.7 & \textbf{20.3} & 18.1 ({\scriptsize 
\textcolor{blue}{+0.4}}) \\ 
Slot Detection & Massive {\scriptsize (15 \underline{low-resource} languages)} &  \textbf{33.3} & 5.8 & 12.1 & 19.0 ({\scriptsize \textcolor{blue}{+13.1}})\\ 
Rel. Extraction & REDFM {\scriptsize (7 languages)} & \textbf{19.1} & 15.5 & 16.8 & 16.2 ({\scriptsize \textcolor{blue}{+0.7}}) \\ 
\midrule 
\multirow{4}{*}{Average}& Seen & 33.7 & 58.8 & 60.0 & \textbf{61.8} ({\scriptsize \textcolor{blue}{+3.0}}) \\ 
& Unseen & \textbf{39.5} & 21.6 & 28.1 & 29.5 ({\scriptsize \textcolor{blue}{+8.0}}) \\ 
& English-only & 55.2 & 58.6 & \textbf{60.3} & 59.3 ({\scriptsize \textcolor{blue}{+0.7}}) \\ 
& All  & 36.6 & 40.2 & 44.1 & \textbf{45.7} ({\scriptsize \textcolor{blue}{+5.5}}) \\ 
\bottomrule
\end{tabularx}
\caption{\textbf{Zero-shot cross-lingual transfer} performance (F1 score). The table compiles all the seen label schema and unseen label schema evaluation results. Blue numbers highlight the performance improvements over GoLLIE-7B ($\Delta$). Full results for each language can be found in Appendix. (2024/6/13: fixed data processing bugs when reporting results for ACE/REDFM.)}
\label{table:main_results}
\end{table}

\begin{figure*}[ht!]
\centering
\includegraphics[width=\textwidth]{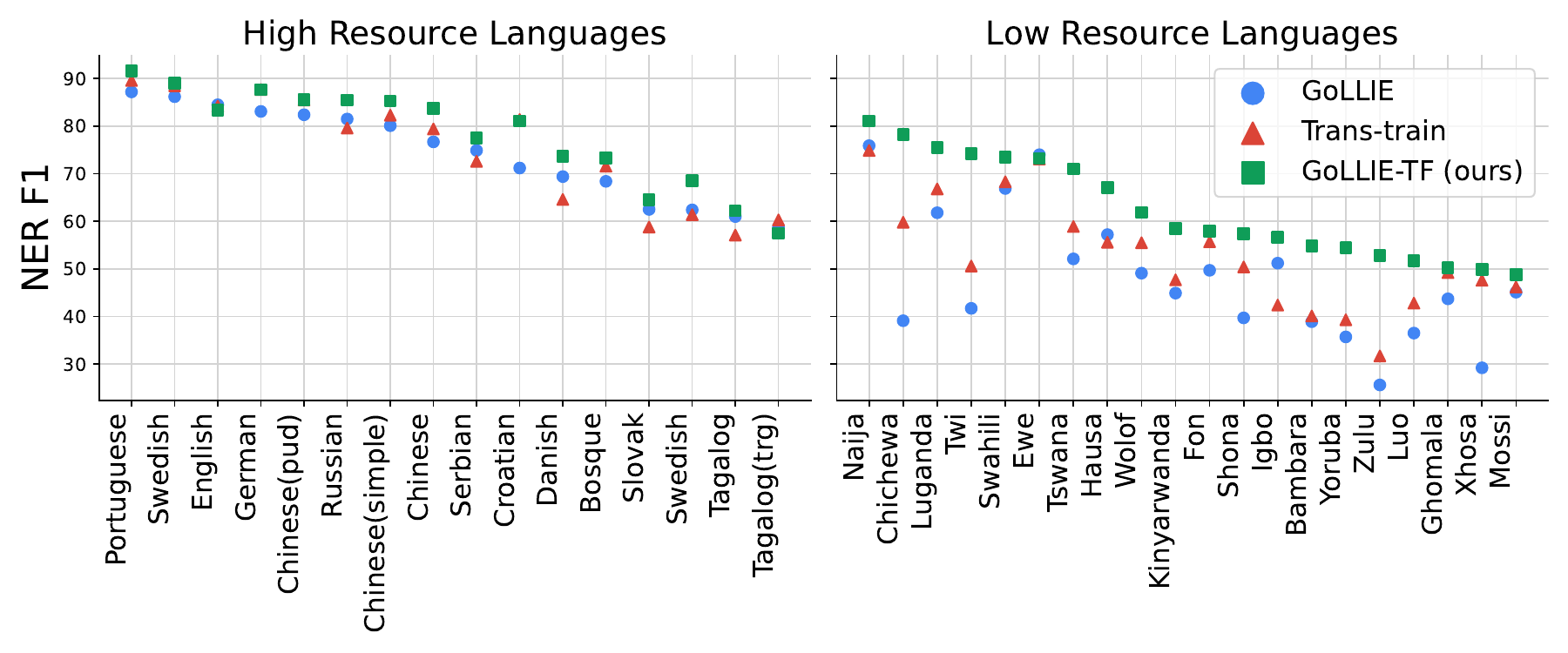}
\vspace{-7mm}
\caption{TransFusion leads to larger NER F1 improvements for low resource languages in MasakhaNER2 (right) compared to high resource languages in UNER (left).}
\label{fig:lang_ner}
\end{figure*}

\section{Results}

We present zero-shot cross-lingual transfer results for IE tasks in Table~\ref{table:main_results}, evaluating both seen and unseen label schemas across 36 languages. Our proposed GoLLIE-TF model consistently outperforms the original GoLLIE, achieving an average F1 score improvement of +4.6 across 11 datasets. Notably, GoLLIE-TF demonstrates significant performance gains in low-resource language NER while mainting English performance on average. For instance, on the MasakhaNER2 dataset, TransFusion boosts F1 from 47.9 to 62.4, surpassing both GPT-4 and the translate-train baseline. Furthermore, GoLLIE-TF supports generalization to unseen label schemas. In particular, TransFusion significantly improves performance on MultiCoNER2 (+12.2), xSID (+20.4), and on low-resource language dataset Massive (+13.1) over GoLLIE, showcasing its adaptability to unseen tasks. 
While GPT-4 still demonstrates superior performance on unseen label schemas, we would like to highlight that our experiments are conducted in a controlled setting. In contrast, for proprietary models, we are unaware of the dataset used, leading to potential dataset contamination.

\paragraph{TransFusion performance on High vs. Low-resource languages.}  Figure~\ref{fig:lang_ner} reveals a noteworthy trend: GoLLIE-TF exhibits substantial performance enhancements particularly in low-resource language settings. This underscores the significance of leveraging external Machine Translation systems to enrich input data for such languages. We followed the categorization of high and low-resource languages from \citet{nllb2022}, which categorizes a language as low-resource if there are fewer than 1M publicly available deduplicated bitext samples. 
While the performance disparity between GoLLIE-TF and other models remains modest in high-resource language scenarios, a notable performance gap emerges in the low-resource language domain. Furthermore, results on the unseen-label low-resource language dataset, Massive, also show that GoLLIE-TF signficiantly outperforms Translate-Train, as shown in  in Table~\ref{table:main_results}.

\begin{wraptable}{r}{0.35\textwidth}
\centering
\vspace{-0.5cm}
\small
\begin{tabular}{ll}
\toprule
Model & MasakhaNER2 F1\\
\midrule
GoLLIE-TF & 62.4\\
- w/o annotate & 55.7 (\textcolor{blue}{-10.5})\\
\bottomrule
\end{tabular}
\caption{Ablation study.}
\vspace{-0.5cm}
\label{table:ablation}
\end{wraptable}

\subsection{Ablation Study}

\paragraph{Analyzing Performance Improvements}
Table~\ref{table:ablation} shows a critical insight into the performance gains observed in the TransFusion framework, particularly in the \textit{annotate} step on the English translation, which plays a crucial role in enhancing the performance of MasakhaNER2.
We conduct an ablation study wherein we trained a variant of GoLLIE-TF, termed GoLLIE-TF (w/o \textit{annotate}), directly generating predictions on target language text from the unlabelled English text. We observe a notable performance drop from 62.4 to 55.7 F1 score. This observation underscores the significance of TransFusion's ability to leverage English annotations during test time, resulting in more precise predictions.

\paragraph{Robustness to translation quality.}

TransFusion offers a distinct advantage by leveraging an external multilingual MT system to augment its dataset with English translations. 
However, the efficacy of this approach hinges on the translation quality provided by the external MT system. 
\begin{wrapfigure}{r}{0.35\textwidth}
\centering
\includegraphics[width=0.35\textwidth]{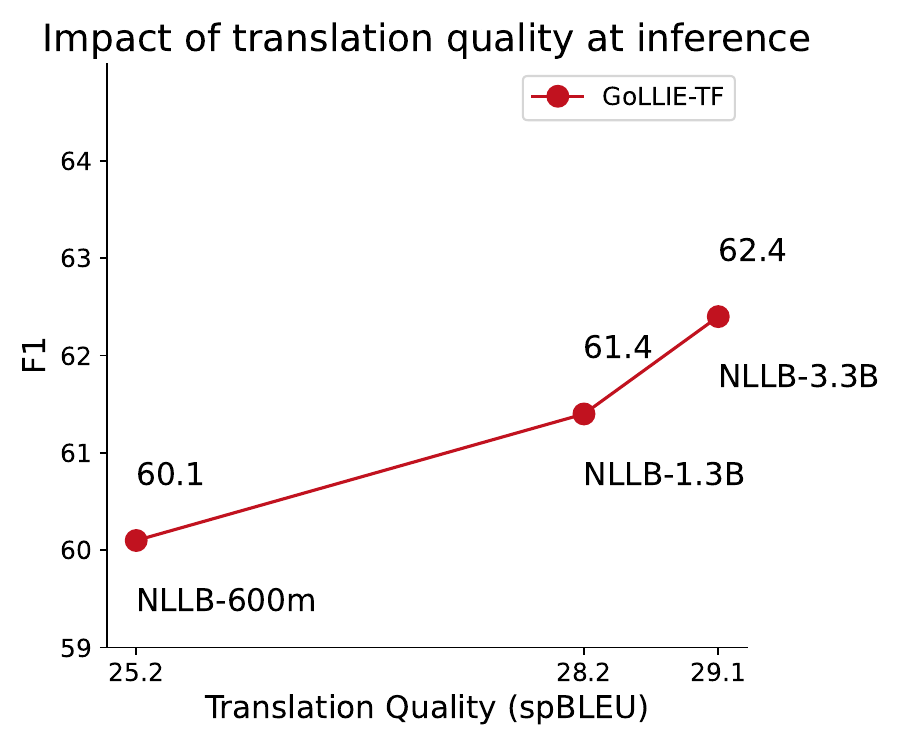}
\caption{TransFusion robustness to different translation systems.} 
\vspace{-1cm}
\label{fig:trans_quality}
\end{wrapfigure}

In Figure~\ref{fig:trans_quality}, we explore this aspect by evaluating GoLLIE-TF's performance with three different MT systems (NLLB-200-{600m, 1.3b, 3.3b}) and use Flores-200 translation benchmark (\texttt{X} to English)~\citep{nllb2022} to measure translation quality (spBLUE) of languages covered by MasakhaNER2. Our experiments reveal that GoLLIE-TF exhibits robustness across various MT systems, as we observe that the F1 score on MasakhaNER2 does not exhibit a significant drop, however performance does improve with a stronger translation system.

\subsection{Enhancing GPT-4 with TransFusion}
\label{sec:gpt4}
\begin{figure*}[t!]
\centering
\includegraphics[width=1.0\linewidth]{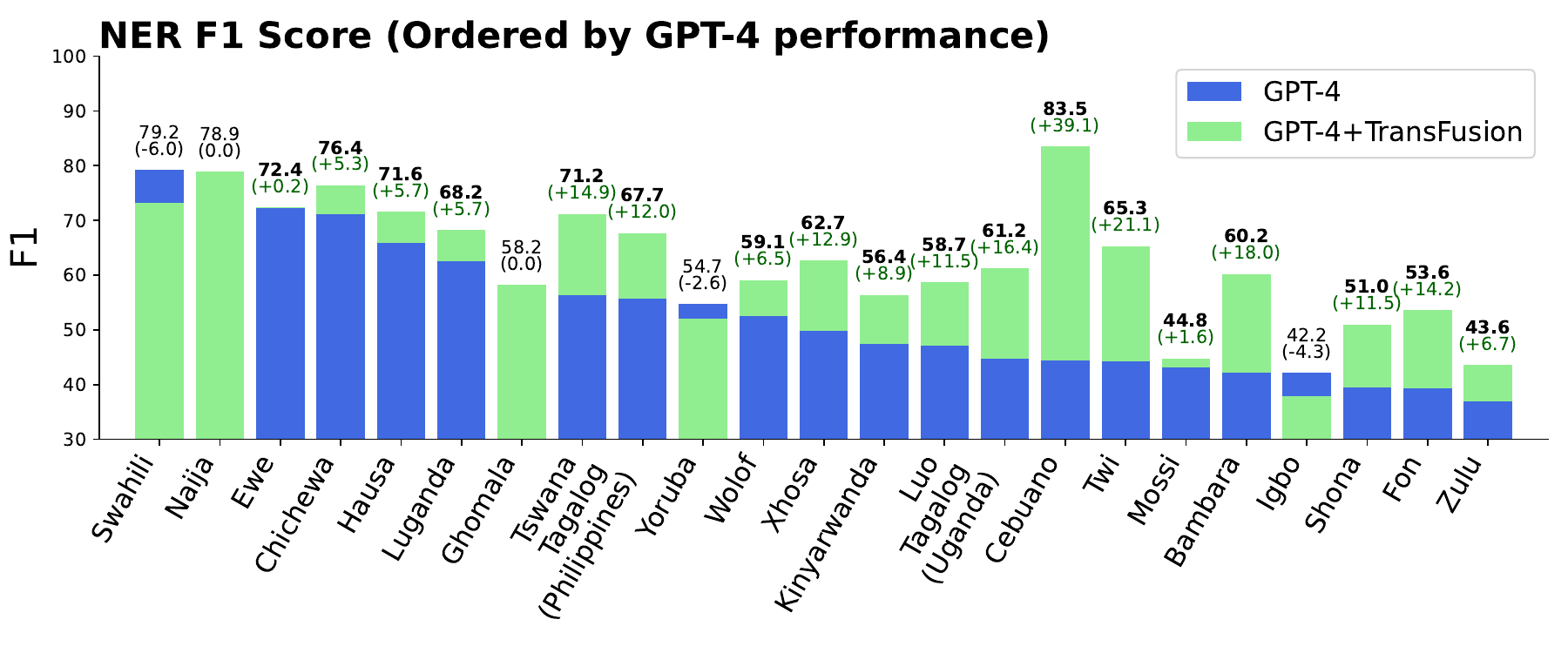}
\caption{GPT-4 + TransFusion framework improves NER on low-resource language from MasakhaNER2 and UNER subsets. 
On average, GPT-4 + TransFusion improves average F1 from 53.4 to 62.}
\label{fig:gpt-4}
\vspace{-5mm}
\end{figure*}

Despite GPT-4's pre-training on multilingual corpora, a notable performance gap persists between its English NER capabilities on CoNLL03 (80 F1) and its performance on low-resource languages (54.2 F1).
In Figure~\ref{fig:gpt-4}, we employ the TransFusion instruction, asking GPT-4 for predictions on the English translation and to then use these labels to predict on the target lagnauge sentence. We show TransFusion prompting yields a substantial enhancement in GPT-4's NER performance across MasakhaNER2 and three additional low-resource languages from the UNER dataset (Cebuano, Tagalog (Philippines), and Uganda), improving the average F1 score from 53.4 to 62. This shows the GPT-4 can follow TransFusion prompting framework to leverage its English predictions to make more accurate predictions on low-resource languages.

\subsection{TransFusion with Encoder-only Models}
\label{sec:encoder}

\begin{wraptable}{r}{0.4\textwidth}
\centering
\small
\vspace{-7mm}
\begin{tabular}{lc}
\toprule
Model & F1\\
\midrule
EasyProject~\citep{chen2022frustratingly} & 64.9\\
Codec~\citep{le2024constrained} & 70.1\\
\midrule
AfroXLM-R (550M) & 58.8\\
+ Trans-train & 65.8\\
+ \textbf{TransFusion} (ours) & \textbf{72.1}\\
\bottomrule
\end{tabular}
\caption{F1 of encoder-only multilingual LM on MasakhaNER2, average of 3 random seeds.}
\label{tab:encoder}
\vspace{-7mm}
\end{wraptable}

We have demonstrated that TransFusion can be applied to GPT-4 to improve low-resource language NER performance and also with the decoder-only LLM GoLLIE, which has the benefit of generalizing to unseen label schemas. In this section, we experiment with encoder-only multilingual LMs~\citep{devlin2019mbert} as the encoder architecture is one of the standard approaches for NER tasks used in practice.

As encoder-only models generally assume the same label schema between fine-tuning and evaluation, we focus on the seen label schema experiment setting, where we use CoNLL03 English as training data and test on the full test set of MasakhaNER2. We use AfroXLM-R~\citep{alabi-etal-2022-adapting}, an African language pre-trained language model as MasakhaNER is an African language dataset. 
For each language, we fine-tuned the model on a combination (50/50\%) of English and translation (Trans-train) or TransFusion data for 5 epochs with a learning rate of 2e-5. The specific TransFusion implementation is introduced in Section~\ref{sec:fusion}.

In Table~\ref{tab:encoder}, we show the effectiveness of the TransFusion framework which boosts the F1 from 58.8 to 72.1 F1 on MasakhaNER2 with AfroXLM-R. In addition, it outperforms the Trans-train baseline significantly with a +6.3 F1 improvement and achieves state-of-the-art performance on MasakhaNER2, surpassing the previous state-of-the-art Codec~\citep{le2024constrained}. 
Codec uses constrained decoding within a translation model to generate precise label projections from English to the target language for Translate-test. In contrast, TransFusion introduces a model that learns to fuse annotations, showing robustness to errors in English annotation predictions.
Overall, this shows the generalization of the TransFusion to the encoder-only multilingual language model.

\subsection{Error Analysis}
\begin{figure*}[t!]
\centering
\includegraphics[width=1.0\linewidth]{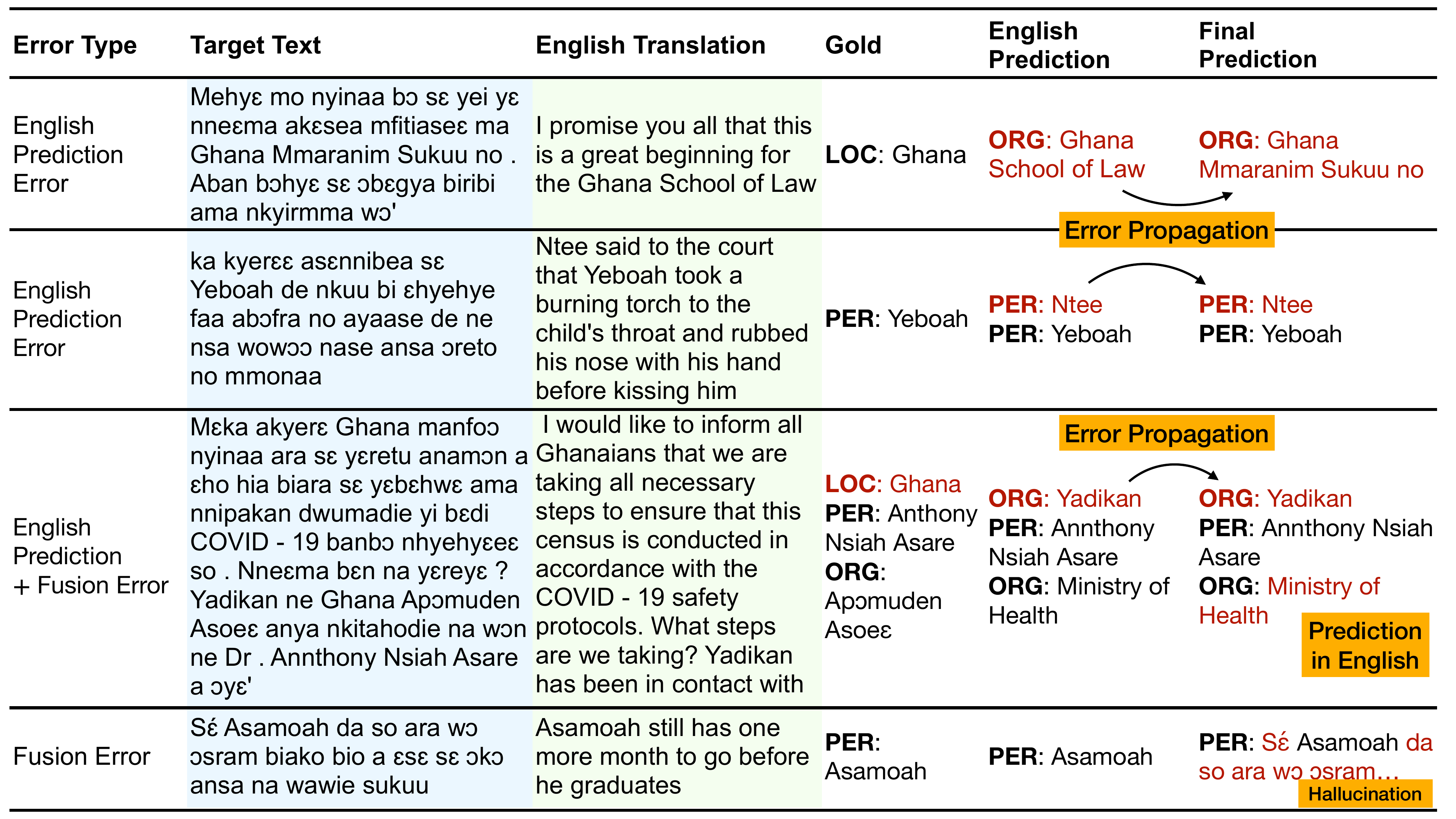}
\caption{\textbf{Error analysis} of GoLLIE-TF's 31 incorrect predictions on MasakhaNER2 (Akan). 
Two common errors are categorized as English prediction error (22/31) and fusion error (12/31).}
\label{fig:error}
\end{figure*}

To understand the reasons why GoLLIE-TF makes mistakes, we conducted a manual error analysis on the MasakhaNER2 (Akan) subset and annotated 31 errors from the model. In Figure~\ref{fig:error}, we show examples of two common error types made by GoLLIE-TF: (1) English prediction errors, where the predictions on English translation are incorrect, and (2) Fusion errors, where the error arises from the fusion stage.
We identified 22 out of 31 cases where the model made errors in predicting NER for the English translation text, and thus these errors propagated to the final predictions. On the other hand, we found 12 out of 31 cases where the model made incorrect fusion processes, leading to hallucinations in the final predictions or predictions in the English text. 

\section{Related Work}

\paragraph{Multilingual language models.}
Multilingual language models~\citep{devlin2019mbert,conneau2019xlm,conneau2019xlmr,xue2021mt5,scao2022bloom,asai2023buffet}, have facilitated cross-lingual transfer by leveraging pre-training on large-scale multilingual corpora.  Recent models such as Gemini~\citep{team2023gemini} show emergent capabilities such as ultra low-resource language translation with a book and wordlist in context.
However, their performance tends to be subpar on languages that were not seen during pre-training or are underrepresented in the training data~\citep{adelani2021masakhaner, ebrahimi2021americasnli}. To address this limitation, several approaches have been explored, including bilingual models~\citep{lan2020focused,wang-etal-2020-extending}, language-specific extensions~\citep{ogueji-etal-2021-small, alabi-etal-2022-adapting, yoon2024langbridge}, continued training~\citep{wang-etal-2020-extending,pfeiffer2020unks,wang2022expanding,imani2023glot500}. Recently, multilingual instruction-tuning datasets such as Aya~\citep{singh2024aya,ustun2024aya} focusing on text generation and IEPile~\citep{gui2024iepile} (English and Chinese) have been proposed to facilitate this direction of research.

\paragraph{Translation for cross-lingual transfer.}
To enhance LLM on multilingual NLP tasks, translating train or test data into English has proven as an effective approach~\citep{hu2020xtreme,xue2021mt5}. Recent studies on prompting LLMs with translation demonstrate improvements on multilingual math reasoning~\citep{shi2022language}, text generation~\citep{huang2023not,intrator2024breaking,liu2024translation} and sentence classification~\citep{etxaniz2023multilingual}. In contrast, our work focuses on challenging IE tasks that require extracting span annotations on the target language directly, instead of generating text. It is even more challenging to construct translated data for translate-train as span annotations are missing after translation.
To solve this, word alignment models~\citep{giza,dyer2013fast,lan2021crfalign,dou2021awesome,parekh2023contextual,le2024constrained} and a simple mark-then-translate approach~ \citep{lee2018qaquote, lewis2020mlqa, hu2020xtreme, bornea2021multilingual,chen2022frustratingly} have been utilized to project labels across different languages. In contrast, we train a model to fuse annotations from English and directly make predictions on target language.

\section{Conclusion}
We introduce TransFusion, a framework that bridges the performance gap between high and low-resource languages in information extraction by leveraging machine translation. Our experiments demonstrate that TransFusion significantly improves the zero-shot cross-lingual transfer capabilities of instruction-tuned LLMs, surpassing both proprietary models and encoder-only architectures on low-resource languages NER. This work demonstrates the potential of translation-based techniques to unlock the power of LLMs for a wider range of low-resource languages, paving the way for more inclusive and equitable IE capabilities across diverse linguistic communities.

\section*{Limitation and broader impact}
The NER experiments conducted on GPT-4 have yielded promising results for low-resource languages.
However, concerns remain regarding potential data contamination resulting from the possibility that GPT-4 was pre-trained or fine-tuned on the test data.\footnote{https://hitz-zentroa.github.io/lm-contamination/blog/}
The Translation-and-fusion framework, while effective in enhancing cross-lingual transfer, does introduce additional inference costs during test time inference. These additional steps include translation using an external MT system and annotation processes, which can contribute to an increased number of token generation. This is similar to chain-of-thought prompting or retrieval augmented generation, which uses additional computational cost at inference for better quality generation.
Thus, practitioners should consider the trade-off between performance and efficiency when deciding to adopt the Translation-and-fusion approach. 

The proposed method carries minimal risk, given that it addresses a traditional IE task. Its primary objective is to enhance IE cross-lingual transfer performance for low-resource languages lacking annotated training data. Consequently, our work aims to have a broader impact by facilitating research for global communities with diverse languages.

\section*{Acknowledgements}
This material is based upon work supported by the NSF (IIS-2052498) and IARPA via the BETTER and HIATUS programs (2019-19051600004, 2022-22072200004). The views and conclusions contained herein are those of the authors and should not be interpreted as necessarily representing the official policies, either expressed or implied, of NSF, ODNI, IARPA, or the U.S. Government. The U.S. Government is authorized to reproduce and distribute reprints for governmental purposes notwithstanding any copyright annotation therein.

\clearpage
\newpage

\bibliographystyle{plainnat}
\bibliography{custom}
\clearpage

\newpage
\appendix
\section{Appendix}

\begin{table}[ht]
\centering
\small
\begin{tabularx}{\textwidth}{lX}
\toprule
\textbf{Dataset} & \textbf{Language Code} \\
\midrule
MasakhaNER2.0~\citep{adelani2022masakhaner} & Bambara (bam),  Ghomala (bbj), Ewe (ewe), Fon (fon), Hausa (hau), \\ 
afl-3.0 License& Igbo (ibo), Kinyarwanda (kin), Luganda (lug), Luo (luo),  Mossi (mos),  \\
  \url{masakhane/masakhaner2}               &  Nyanja (nya),  Naija (pcm),  Shona (sna),  Swahili (swh),  Tswana (tsn)\\
                  & Twi (twi),  Wolof (wol),  Xhosa (xho),  Yoruba (yor), Zulu (zul) \\
                 \midrule
UNER~\citep{mayhew2023uner}& Cebuano (ceb\_gja),  Danish (da\_ddt), German (de\_pud),  \\ 
       \url{universalner.org/} & English (en\_ewt), English (en\_pud), Croatian (hr\_set),  \\
       (Unknown License)  & Portuguese (pt\_bosque), Portuguese (pt\_pud), 
         Russian (ru\_pud),   \\ 
         & Slovak (sk\_snk),  Serbian (sr\_set), \\
        & Swedish (sv\_pud),  
        Swedish (sv\_talbanken),  \\
        & Tagalog (tl\_trg), Tagalog (tl\_ugnayan),  Chinese (zh\_gsd), \\
        & Chinese (zh\_gsdsimp),  
        Chinese (zh\_pud)\\
        \midrule
ACE05~\citep{walker2006ace}& English (en), Arabic (ar), Chinese (zh) \\
LDC license: LDC2006T06 & \\
\midrule
MultiNERD~\citep{tedeschi-navigli-2022-multinerd} & German (de), Spanish (es), French (fr), Italian (it), Dutch (nl), \\
            CC BY-NC-SA 4.0 &  Polish (pl), Portuguese (pt), Russian (ru), Chinese (zh)\\ \url{Babelscape/multinerd}\\
            \midrule
MultiCoNER2~\citep{fetahu2023multiconer} & Bengali (bn), German (de), Spanish (es), Persian (fa), French (fr), \\ 
CC BY 4.0  & Hindi (hi), Italian (it), Portuguese (pt), Swedish (sv), \\
             \url{MultiCoNER/multiconer_v2} & Ukrainian (uk), Chinese (zh), English (en) \\
             \midrule
xSID~\citep{van-der-goot-etal-2021-xsid} & Arabic (ar), Danish (da), German (de), 
English (en), Indonesian (id),  \\
        CC BY-SA 4.0 & Italian (it), Japanese (ja), Kazakh (kk), Dutch (nl), Serbian (sr), \\
         & Turkish (tr), Chinese (zh) \\
        \midrule
MultiTO~\citep{Schuster2018CrosslingualTL} & English (en), Spanish (es), Thai (th) \\
CC-BY-SA& \\ \midrule
RED-FM~\citep{cabot2023redfm}  & Arabic (ar), German (de), English (en), Spanish (es), French (fr),  \\
CC BY-SA 4.0 &Italian (it), Chinese (zh) \\
\url{Babelscape/REDFM} & \\
\midrule
MASSIVE~\citep{fitzgerald2022massive} &  Afrikaans (af-za),
Amharic (am-et),
Azeri (az-za),
Bengali (bn-bd),\\
CC BY 4.0&Armenian (hy-am),
Georgian (ka-ge),Khmer (km-kh),
Mongolian (mn-mn),\\
\url{AmazonScience/massive}&Burmese (my-mm),
Kannada (kn-in),
Malayalam (ml-in),\\
& Tamil (ta-in),
Telugu (te-in),
Tagalog (tl-ph),
Welsh (cy-gb)\\
\bottomrule
\end{tabularx}
\caption{Evaluation datasets used and the language code for each dataset.}
\end{table}

\begin{figure*}[htp!]
\centering
\includegraphics[width=1.05\linewidth]{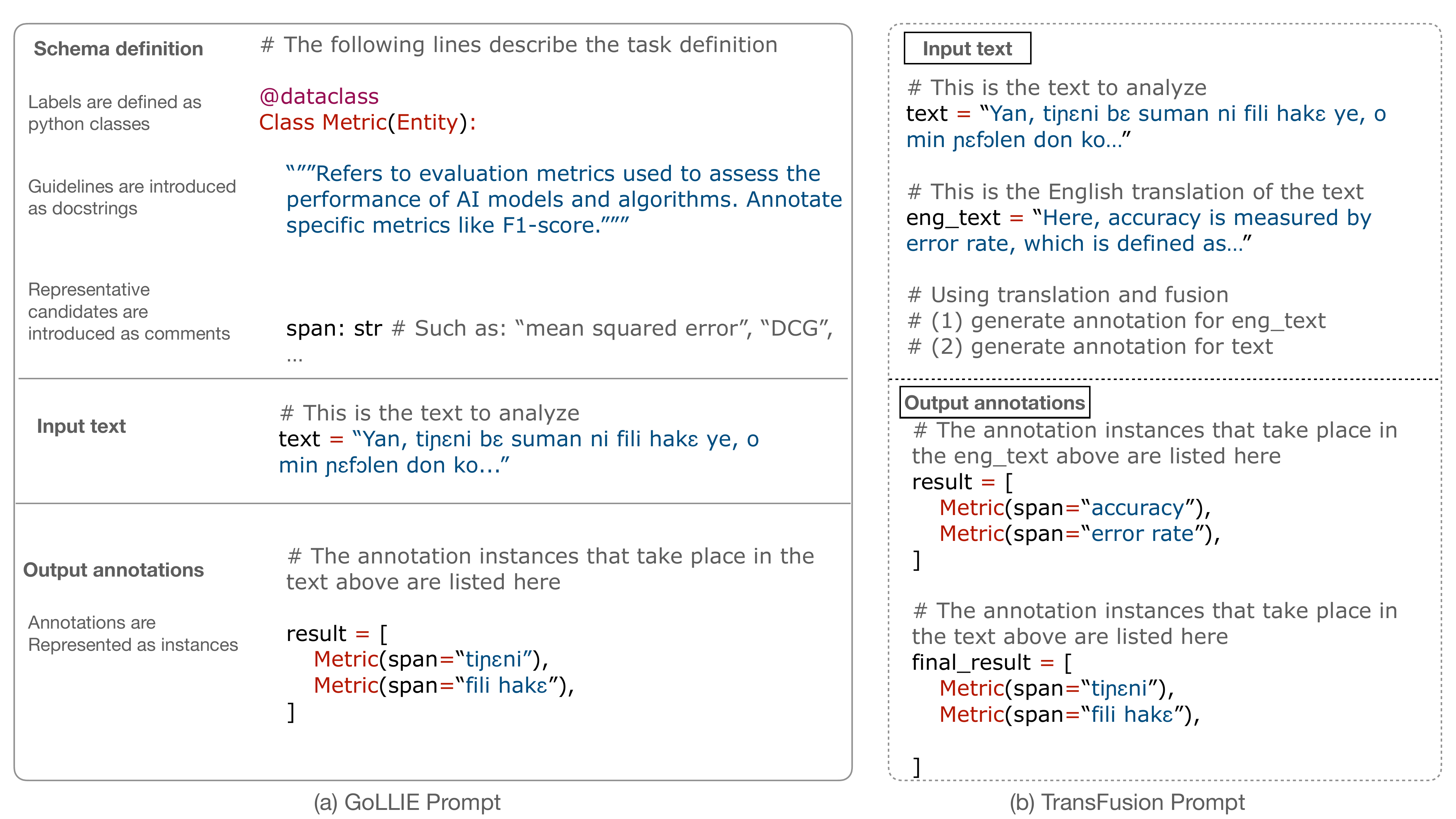}
\caption{Example of input and output representation. (left) An example of a named entity recognition prompt and output annotations. (right) The same example but with translation text appended in the input prompt with instructions to guide the model to generate annotations on English translation text first, followed by annotations on the target language. 
}
\label{fig:x-gollie}
\end{figure*}

\begin{figure}[ht!]
\centering
\includegraphics[width=\textwidth]{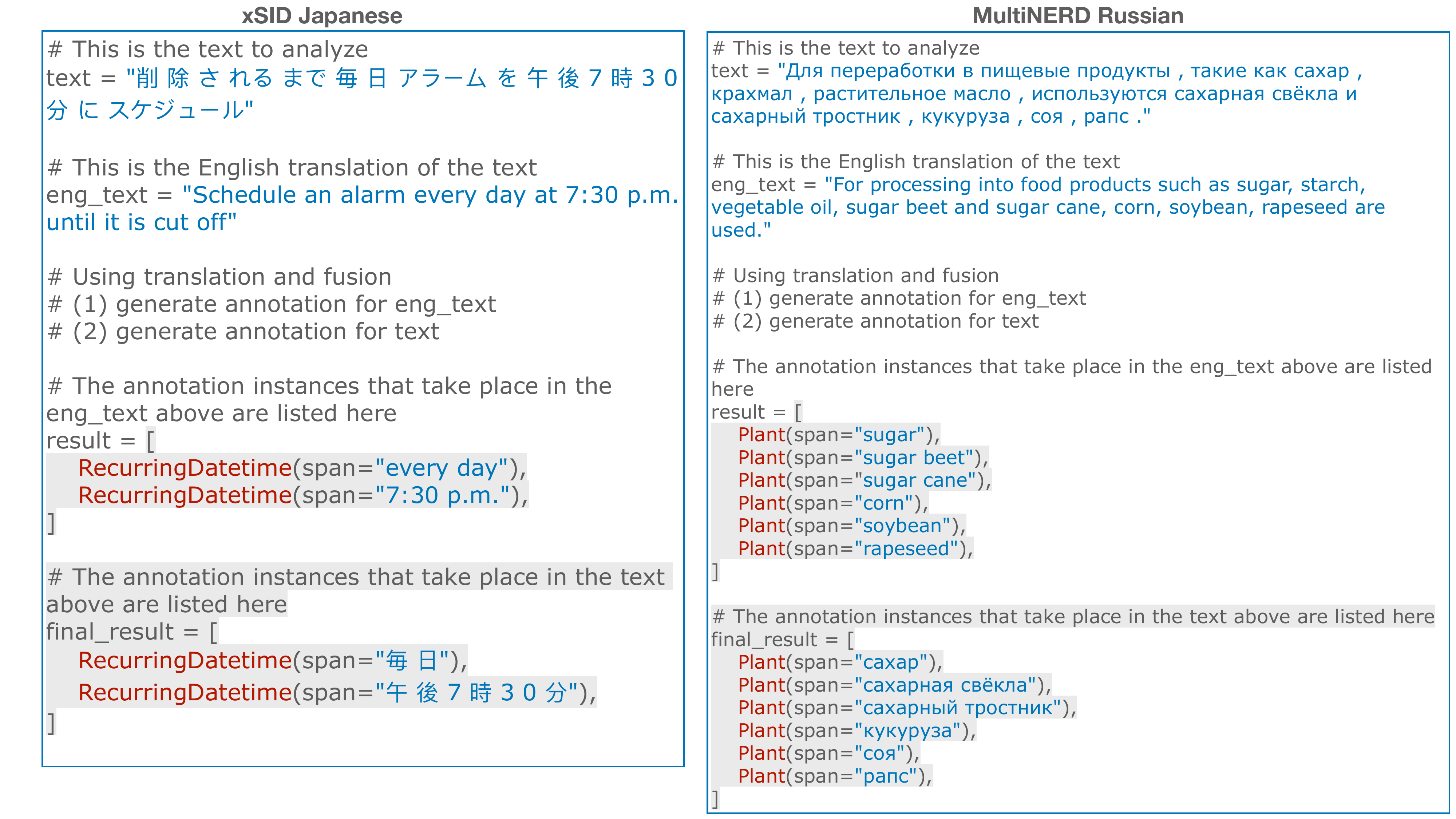}
\caption{Examples of GoLLIE-TF model generation out (colored in gray).}
\label{fig:transfusion_stats}
\end{figure}

\begin{figure}[ht!]
\centering
\includegraphics[width=0.7\textwidth]{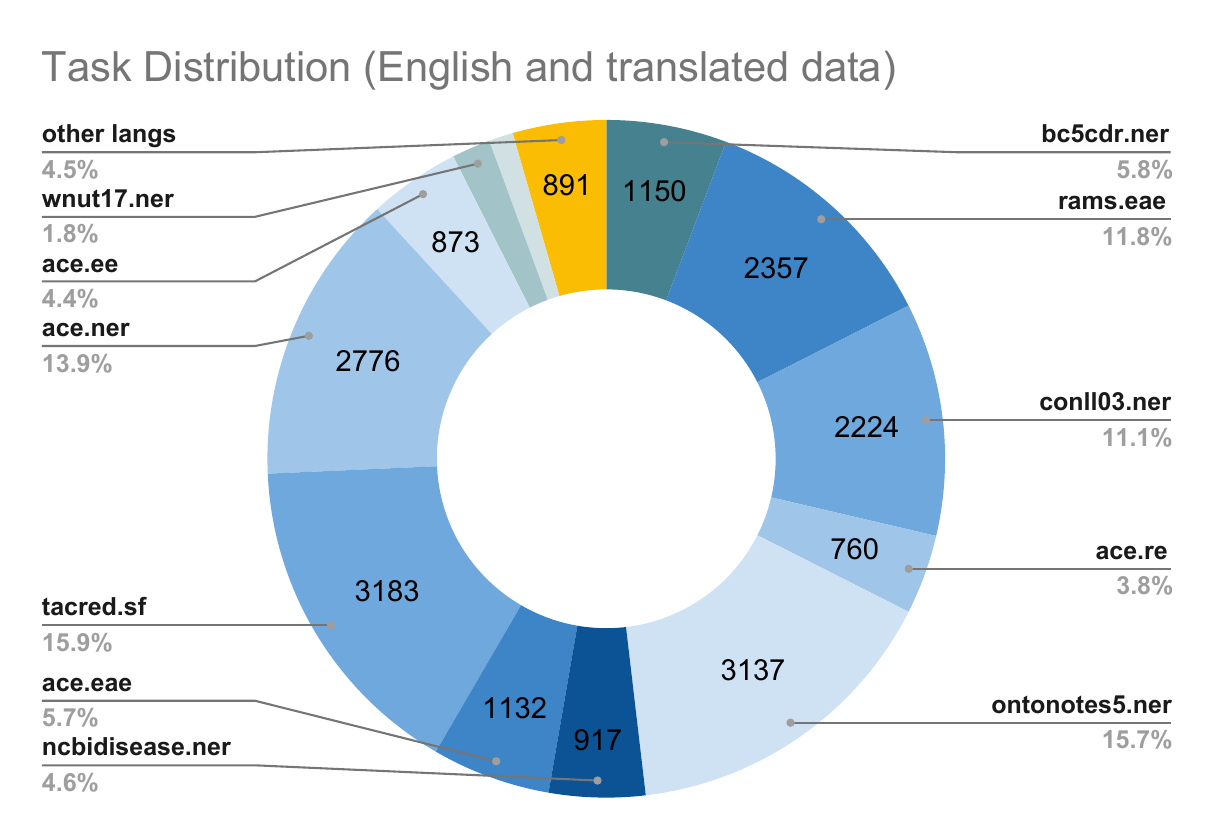}
\includegraphics[width=0.7\textwidth]{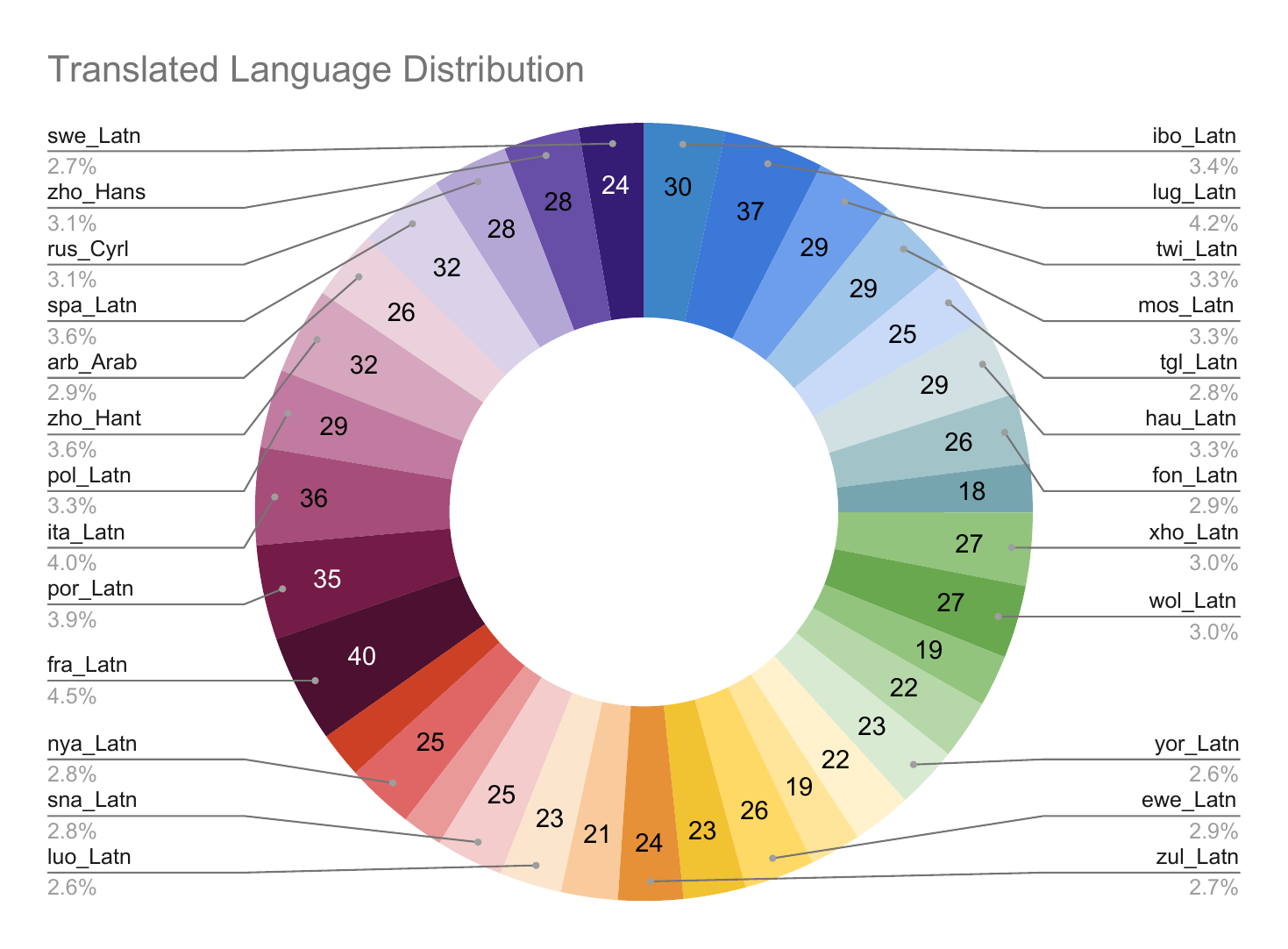}
\includegraphics[width=0.7\textwidth]{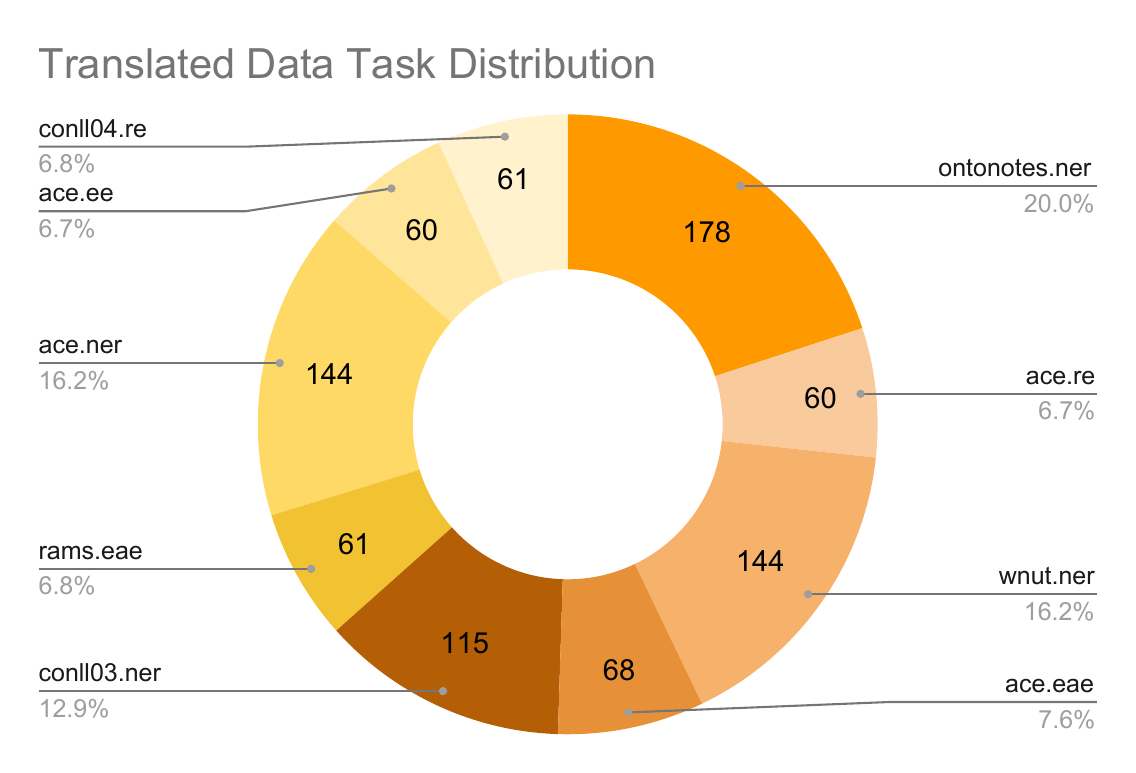}
\caption{TransFusion training dataset mixture for a total of 20,000.}
\label{fig:transfusion_stats}
\end{figure}

\begin{table}[ht]
    \centering
    \small
    \begin{tabularx}{0.8\textwidth}{lXXXXX}
    \toprule
    Dataset & Seed 0 & Seed 1 & Seed 2 & Mean &Std dev\\
    \midrule
    masakhaner.bam.ner & 54.8 & 53.7 & 56.1 & 54.9 & 1.2 \\
masakhaner.bbj.ner & 50.2 & 46.2 & 50.9 & 49.1 & 2.6 \\
masakhaner.ewe.ner & 73.2 & 72.7 & 73.1 & 73.0 & 0.3 \\
masakhaner.fon.ner & 57.9 & 54.3 & 55.7 & 56.0 & 1.8 \\
masakhaner.hau.ner & 67.1 & 65.6 & 66.2 & 66.3 & 0.8 \\
masakhaner.ibo.ner & 56.6 & 54.2 & 55.7 & 55.5 & 1.3 \\
masakhaner.kin.ner & 58.5 & 59.5 & 59.6 & 59.2 & 0.6 \\
masakhaner.lug.ner & 75.5 & 74.5 & 75.1 & 75.0 & 0.5 \\
masakhaner.luo.ner & 51.7 & 51.6 & 51.5 & 51.6 & 0.1 \\
masakhaner.mos.ner & 48.8 & 43.8 & 44.4 & 45.7 & 2.7 \\
masakhaner.nya.ner & 78.2 & 78.7 & 78.9 & 78.6 & 0.3 \\
masakhaner.pcm.ner & 81.1 & 80.8 & 80.6 & 80.8 & 0.2 \\
masakhaner.sna.ner & 57.4 & 59.2 & 56.7 & 57.7 & 1.3 \\
masakhaner.swh.ner & 73.5 & 72.6 & 72.9 & 73.0 & 0.5 \\
masakhaner.tsn.ner & 71.0 & 70.3 & 71.1 & 70.8 & 0.5 \\
masakhaner.twi.ner & 74.2 & 68.6 & 76.6 & 73.1 & 4.1 \\
masakhaner.wol.ner & 61.9 & 55.6 & 60.2 & 59.2 & 3.2 \\
masakhaner.xho.ner & 49.9 & 54.4 & 51.3 & 51.9 & 2.3 \\
masakhaner.yor.ner & 54.4 & 52.4 & 53.4 & 53.4 & 1.0 \\
masakhaner.zul.ner & 52.8 & 53.3 & 51.4 & 52.5 & 1.0 \\
Average & 62.4 & 61.1 & 62.1 & 61.9 & 0.7 \\
\midrule
massive.en-us.ner & 53.6 & 51.6 & 51.6 & 52.3 & 1.1 \\
massive.af-za.ner & 24.2 & 21.2 & 24.2 & 23.2 & 1.7 \\
massive.am-et.ner & 6.5 & 5.4 & 7.2 & 6.4 & 0.9 \\
massive.az-az.ner & 1.2 & 1.3 & 1.3 & 1.2 & 0.1 \\
massive.bn-bd.ner & 18.1 & 18.8 & 19.4 & 18.8 & 0.6 \\
massive.hy-am.ner & 19.4 & 16.2 & 21.1 & 18.9 & 2.5 \\
massive.ka-ge.ner & 18.4 & 16.0 & 19.6 & 18.0 & 1.9 \\
massive.km-kh.ner & 20.4 & 21.1 & 23.2 & 21.5 & 1.5 \\
massive.mn-mn.ner & 5.8 & 5.4 & 5.2 & 5.5 & 0.3 \\
massive.my-mm.ner & 31.7 & 32.4 & 33.2 & 32.4 & 0.8 \\
massive.kn-in.ner & 17.2 & 14.2 & 20.7 & 17.3 & 3.2 \\
massive.ml-in.ner & 11.0 & 10.6 & 10.3 & 10.7 & 0.4 \\
massive.ta-in.ner & 17.0 & 11.6 & 17.3 & 15.3 & 3.2 \\
massive.te-in.ner & 18.8 & 17.6 & 23.5 & 20.0 & 3.1 \\
massive.tl-ph.ner & 32.0 & 32.0 & 34.7 & 32.9 & 1.5 \\
massive.cy-gb.ner & 8.3 & 5.8 & 7.0 & 7.0 & 1.2 \\
Average & 19.0 & 17.6 & 20.0 & 18.8 & 1.2 \\
\bottomrule
    \end{tabularx}
    \caption{We report GoLLIE-TF on MasakhaNER2 and Massive for 3 different seeds.}
\end{table}

\begin{table}
\centering
\small
\begin{tabular}{l|cccc}

\toprule
 & GPT-4 & GoLLIE & Trans-train & GoLLIE-TF (ours) \\
\midrule
uner.ceb\_gja.ner & 44.4 & 49.6 & 52.9 & 87.5\\
uner.da\_ddt.ner & 77.2 & 76.7 & 79.4 & 84.8\\
uner.de\_pud.ner & 80.3 & 80.1 & 82.3 & 83.8\\
uner.en\_ewt.ner & 59.9 & 84.7 & 67.6 & 66.4\\
uner.en\_pud.ner & 75.4 & 82.4 & 85.5 & 84.9\\
uner.hr\_set.ner & 82.1 & 83.0 & 87.7 & 89.6\\
uner.pt\_bosque.ner & 82.7 & 84.5 & 84.2 & 81.3\\
uner.pt\_pud.ner & 80.5 & 87.2 & 89.6 & 90.3\\
uner.ru\_pud.ner & 69.8 & 68.3 & 71.6 & 73.3\\
uner.sk\_snk.ner & 70.9 & 71.2 & 81.4 & 85.5\\
uner.sr\_set.ner & 85.9 & 86.2 & 88.5 & 88.9\\
uner.sv\_pud.ner & 73.7 & 81.5 & 79.6 & 85.7\\
uner.sv\_talbanken.ner & 68.7 & 69.4 & 64.6 & 75.7\\
uner.tl\_trg.ner & 55.7 & 58.8 & 60.3 & 54.2\\
uner.tl\_ugnayan.ner & 44.8 & 61.0 & 57.1 & 74.2\\
uner.zh\_gsd.ner & 60.6 & 62.5 & 58.8 & 67.6\\
uner.zh\_gsdsimp.ner & 57.9 & 62.4 & 61.4 & 68.8\\
uner.zh\_pud.ner & 72.0 & 74.8 & 72.6 & 77.7\\
average & 69.0 & 73.6 & 73.6 & 78.9\\
\midrule
ace.en.eae &  24.5 & 97.3 & 97.9 & 98.3\\
multiace.ar.eae & 1.6 & 84.3 & 83.8 & 81.8\\
multiace.zh.eae &  9.6& 96.6 & 97.1 & 77.9\\
average & 11.7  & 92.7 & 92.9 & 86.0\\
\midrule
 ace.en.ee & 27.8 & 67.5 & 64.0 & 60.4\\
 multiace.ar.ee & 24.4 & 16.1 & 12.8 & 25.0\\
 multiace.zh.ee & 11.6 & 44.2 & 43.3 & 46.7\\
average & 21.3 & 42.6 & 40.0 & 44.0\\
\midrule
ace.en.ner & 58.0 & 78.3 & 87.3 & 86.5\\
multiace.ar.ner & 32.3 & 29.5 & 30.3 & 37.5\\
multiace.zh.ner & 34.6 & 68.2 & 66.0 & 60.6\\
average & 41.6 & 58.7 & 61.2 & 61.5\\
\midrule
ace.en.re	& 5.40&	58.2&	59.8&	58.1\\
multiace.ar.re &	3.2 &	14.1&	13.5&	15.8\\
multiace.zh.re &	5.1&	39.5 &	44.8&	43.3\\
average	& 4.6 &	37.3&	39.4&	39.1\\
\midrule
multinerd.de.ner & 75.8 & 69.3 & 73.2 & 74.4\\
multinerd.es.ner & 69.4 & 72.0 & 68.1 & 69.5\\
multinerd.fr.ner & 71.8 & 71.9 & 74.4 & 72.5\\
multinerd.it.ner & 76.2 & 69.8 & 74.2 & 70.5\\
multinerd.nl.ner & 76.9 & 67.8 & 73.0 & 72.5\\
multinerd.pl.ner & 72.1 & 62.0 & 64.0 & 61.5\\
multinerd.pt.ner & 67.7 & 67.7 & 66.3 & 64.9\\
multinerd.ru.ner & 65.3 & 57.9 & 55.7 & 58.7\\
multinerd.zh.ner & 7.8 & 7.1 & 13.9 & 8.8\\
multinerd.ner & 71.5 & 76.2 & 75.6 & 76.2\\
average & 71.9 & 62.2 & 63.9 & 63.0\\
\bottomrule
\end{tabular}
    \caption{Full experimental results (1) for each dataset and language. Format: [task name].[language code].[task].}
\end{table}

\begin{table}[ht!]
\centering
\small
\begin{tabular}{l|cccc}
\toprule
 & GPT-4 & GoLLIE & Trans-train & GoLLIE-TF (ours) \\
\midrule
multiconer2.bn.ner & 43.9 & 2.7 & 7.9 & 27.6\\
multiconer2.de.ner & 54.4 & 27.3 & 30.8 & 33.1\\
multiconer2.es.ner & 44.8 & 18.1 & 23.9 & 26.1\\
multiconer2.fa.ner & 40.1 & 15.6 & 34.9 & 41.4\\
multiconer2.fr.ner & 54.2 & 29.2 & 32.1 & 34.2\\
multiconer2.hi.ner & 46.9 & 5.0 & 14.8 & 33.5\\
multiconer2.it.ner & 51.1 & 41.4 & 46.0 & 46.5\\
multiconer2.pt.ner & 49.7 & 23.6 & 31.5 & 34.7\\
multiconer2.sv.ner & 52.5 & 14.8 & 16.1 & 19.6\\
multiconer2.uk.ner & 55.9 & 41.1 & 47.7 & 51.7\\
multiconer2.zh.ner & 5.1 & 14.0 & 20.9 & 28.3\\
multiconer2.en.ner & 54.6 & 34.1 & 34.7 & 36.7\\
average & 46.1 & 22.2 & 28.4 & 34.5\\
\midrule
xsid.ar.ner & 53.2 & 0.0 & 29.7 & 28.7\\
xsid.da.ner & 48.1 & 2.7 & 15.5 & 16.0\\
xsid.de.ner & 48.9 & 9.8 & 36.0 & 35.5\\
xsid.en.ner & 63.1 & 28.8 & 38.4 & 37.5\\
xsid.id.ner & 49.4 & 0.7 & 25.6 & 23.2\\
xsid.it.ner & 52.1 & 3.4 & 30.2 & 32.8\\
xsid.ja.ner & 28.1 & 10.1 & 32.8 & 26.5\\
xsid.kk.ner & 34.9 & 0.0 & 0.0 & 2.5\\
xsid.nl.ner & 48.9 & 4.9 & 33.8 & 31.4\\
xsid.sr.ner & 48.7 & 0.0 & 19.4 & 16.8\\
xsid.tr.ner & 40.8 & 0.8 & 20.9 & 22.2\\
xsid.zh.ner & 47.3 & 10.7 & 43.5 & 43.7\\
average & 47.0 & 6.0 & 27.1 & 26.4\\
\midrule
multito.en.ner & 51.1 & 35.3 & 39.0 & 40.3\\
multito.es.ner & 1.4 & 2.5 & 3.0 & 2.3\\
multito.th.ner & 7.3 & 15.4 & 18.9 & 11.8\\
average & 19.9 & 17.7 & 20.3 & 18.1\\
\midrule
redfm.ar.re & 18.3 & 11.6 & 9.0 & 13.9\\
redfm.de.re & 31.0 & 22.3 & 24.8 & 13.1\\
redfm.en.re & 19.9 & 14.8 & 18.6 & 15.7\\
redfm.es.re & 17.4 & 13.8 & 18.6 & 14.4\\
redfm.fr.re & 17.1 & 15.2 & 19.2 & 17.6\\
redfm.it.re & 17.2 & 20.0 & 17.1 & 29.1\\
redfm.zh.re & 12.9 & 10.4 & 10.5 & 9.7\\
average & 19.1 & 15.5 & 16.8 & 16.2\\
\bottomrule
\end{tabular}
\caption{Full experimental results (2) for each dataset and language. Format: [task name].[language code].[task].}
\end{table}

\begin{table}[ht!]
\centering
\small
\begin{tabular}{l|cccc}
\toprule
 & GPT-4 & GoLLIE & Trans-train & GoLLIE-TF (ours) \\
 \midrule
 massive.en-us.ner & 55.2 & 45.9 & 54.7 & 53.6 \\
massive.af-za.ner & 52.6 & 8.2 & 23.4 & 24.2 \\
massive.am-et.ner & 17.0 & 0.0 & 0.8 & 6.5 \\
massive.az-az.ner & 25.7 & 4.0 & 11.0 & 1.2 \\
massive.bn-bd.ner & 33.1 & 5.7 & 13.0 & 18.1 \\
massive.hy-am.ner & 33.6 & 1.2 & 11.9 & 19.4 \\
massive.ka-ge.ner & 32.1 & 10.4 & 12.2 & 18.4 \\
massive.km-kh.ner & 33.9 & 0.0 & 11.3 & 20.4 \\
massive.mn-mn.ner & 19.5 & 0.0 & 5.3 & 5.8 \\
massive.my-mm.ner & 27.9 & 4.8 & 15.2 & 31.7 \\
massive.kn-in.ner & 33.1 & 0.0 & 2.6 & 17.2 \\
massive.ml-in.ner & 25.1 & 0.0 & 4.5 & 11.0 \\
massive.ta-in.ner & 30.7 & 1.2 & 5.0 & 17.0 \\
massive.te-in.ner & 28.7 & 0.0 & 0.0 & 18.8 \\
massive.tl-ph.ner & 50.3 & 12.3 & 20.2 & 32.0 \\
massive.cy-gb.ner & 33.6 & 0.0 & 3.1 & 8.3 \\ 
average & 33.3 & 5.9 & 12.1 & 19.0 \\ 
\bottomrule
\end{tabular}
\caption{Full experimental results (3) for each dataset and language. Format: [task name].[language code].[task].}
\end{table}

\begin{table}[h!]
\centering
\small
\begin{tabular}{lcc}
\toprule
Language & GPT-4 & GPT-4+Transfusion \\
\midrule
MasakhaNER2\\
\midrule
bam & 42.2 & \textbf{60.2} \\
bbj & \textbf{58.2} & 52.9 \\
ewe & 72.2 & \textbf{72.4} \\
fon & 39.4 & \textbf{53.6} \\
hau & 65.9 & \textbf{71.6} \\
ibo & \textbf{42.2} & 37.9 \\
kin & 47.5 & \textbf{56.4} \\
lug & 62.5 & \textbf{68.2} \\
luo & 47.2 & \textbf{58.7} \\
mos & 43.2 & \textbf{44.8} \\
nya & 71.1 & \textbf{76.4} \\
pcm & \textbf{78.9} & 75.7 \\
sna & 39.5 & \textbf{51.0} \\
swh & \textbf{79.2} & 73.2 \\
tsn & 56.3 & \textbf{71.2}\\
twi & 44.2 & \textbf{65.3} \\
wol & 52.6 & \textbf{59.1}\\
xho & 49.8 & \textbf{62.7} \\
yor & \textbf{54.7} & 52.1 \\
zul & 36.9 & \textbf{43.6} \\
\midrule
MasakhaNER2 average & 54.2 & \textbf{59.9} \\
\midrule
\midrule
UNER\\
\midrule
ceb\_gja & 44.4 &  \textbf{83.5}\\
tl\_trg & 55.7 & \textbf{67.7} \\
tl\_ugnayan & 44.8 &  \textbf{61.2}\\
\midrule
All average & 53.4 & \textbf{62.0}\\
\bottomrule
\end{tabular}
\caption{Comparison of GPT-4 and GPT-4+Transfusion.}
\end{table}

\begin{figure}
\begin{lstlisting}
# The following lines describe the task definition
@dataclass
class Location(Entity):
    """Roads (streets, motorways) trajectories regions (villages, towns, cities, provinces, countries, continents,
    dioceses, parishes) structures (bridges, ports, dams) natural locations (mountains, mountain ranges, woods, rivers,
    wells, fields, valleys, gardens, nature reserves, allotments, beaches, national parks) public places..."""

    span: str  # Such as: "U.S.", "Germany", "Britain", "Australia", "England"


@dataclass
class Person(Entity):
    """first, middle and last names of people, animals and fictional characters aliases."""

    span: str  # Such as: "Clinton", "Dole", "Arafat", "Yeltsin", "Lebed"


@dataclass
class Organization(Entity):
    """Companies (press agencies, studios, banks, stock markets, manufacturers, cooperatives) subdivisions of
    companies (newsrooms) brands political movements ..."""

    span: str  # Such as: "Reuters", "U.N.", "NEW YORK", "CHICAGO", "PUK"


# This is the text to analyze
text = "..."

# This is the English translation of the text
eng_text = "..."

# Using translation and fusion
# (1) generate annotation for eng_text
# (2) generate annotation for text

# The annotation instances that take place in the eng_text above are listed here
result = []

# The annotation instances that take place in the text above are listed here
final_result = [
]
\end{lstlisting}
\caption{Annotation guidelines for MasakhaNER and UNER.}
\end{figure}

\begin{figure}
\begin{lstlisting}
# The following lines describe the task definition
class Artist(Entity):
    """A person who creates or practices art, such as a painter, sculptor, musician, or performer."""

    span: str  # Such as: "bob dylan", "paul mccartney", "duke ellington", "william shakespeare", "richard rodgers"


class Athlete(Entity):
    """A person who actively participates in sports or athletic events and competitions."""

    span: str  # Such as: "michael jordan", "kyle larson", "batman", "bob hope", "jeff gordon"

# ... SKIP

class Symptom(Entity):
    """A physical or mental manifestation that is experienced as a consequence of a disease or medical condition."""

    span: (
        str  # Such as: "pain", "hiv/aids", "anxiety", "inflammation", "cystic fibrosis"
    )


class Disease(Entity):
    """A disorder or abnormal condition that affects the body or mind and causes specific symptoms or impairments."""

    span: str  # Such as: "tuberculosis", "pneumonia", "multiple sclerosis", "schizophrenia", "smallpox"


# This is the text to analyze
text = ""

# This is the English translation of the text
eng_text = ""

# Using translation and fusion
# (1) generate annotation for eng_text
# (2) generate annotation for text

# The annotation instances that take place in the eng_text above are listed here
result = []

# The annotation instances that take place in the text above are listed here
final_result = [
]
\end{lstlisting}
\caption{Annotation guidelines for MultiCoNER2.}
\end{figure}

\begin{figure}
\begin{lstlisting}
# The following lines describe the task definition
@dataclass
class Celestial(Entity):
    """Planets, stars, asteroids, comets, nebulae, galaxies and other astronomical objects."""

    span: str  # Such as: "Earth", "Sun", "Jupiter", "Moon", "Milky Way"


@dataclass
class Mythological(Entity):
    """Mythological and religious entities."""

    span: str  # Such as: "Thomas Aquinas", "Zeus", "Satan", "Ra", "Aphrodite"


@dataclass
class Biological(Entity):
    """Genus of fungus, bacteria and protoctists, families of viruses, and other biological entities."""

    span: str  # Such as: "Escherichia coli", "E. coli", "Plasmodium", "Aspergillus", "HIV"

# SKIP...

@dataclass
class Time(Entity):
    """Specific and well-defined time intervals, such as eras, historical periods, centuries, years and important days. No months and days of the week."""

    span: str  # Such as: "Middle Ages", "Great Depression", "Renaissance", "Bronze Age", "half-life"


# This is the text to analyze
text = ""

# This is the English translation of the text
eng_text = ""

# Using translation and fusion
# (1) generate annotation for eng_text
# (2) generate annotation for text

# The annotation instances that take place in the eng_text above are listed here
result = []

# The annotation instances that take place in the text above are listed here
final_result = [
]
\end{lstlisting}
\caption{Annotation guideline for MultiNERD.}
\end{figure}

\begin{figure}
\begin{lstlisting}
# The following lines describe the task definition
@dataclass
class WeatherNoun(Entity):
    """Names or terms related to weather conditions or forecasts."""

    span: str  # Such as: "weather", "temperature", "forecast", "degrees", "degree"

# SKIP...

@dataclass
class Negation(Entity):
    """Words used to negate, disagree or express the opposite of something."""

    span: str  # Such as: "instead", "not"


@dataclass
class TimerAttributes(Entity):
    """Words describing attributes, properties or settings related to a timer."""

    span: str  # Such as: "repeat"


@dataclass
class NewsType(Entity):
    """Words indicating the type or category of news content."""

    span: str  # Such as: "news"


@dataclass
class WeatherTemperatureUnit(Entity):
    """Words specifying the unit used to measure temperature."""

    span: str  # Such as: "celsius", "f", "c", "celcius", "fahrenheit"


# This is the text to analyze
text = ""

# This is the English translation of the text
eng_text = ""

# Using translation and fusion
# (1) generate annotation for eng_text
# (2) generate annotation for text

# The annotation instances that take place in the eng_text above are listed here
result = []

# The annotation instances that take place in the text above are listed here
final_result = [
]
\end{lstlisting}
\caption{Annotation guidelines for Massive.}
\end{figure}

\begin{figure}
\begin{lstlisting}
    # The following lines describe the task definition
@dataclass
class WeatherAttribute(Entity):
    """Words describing attributes related to weather conditions like precipitation, temperature, etc."""

    span: str  # Such as: "rain", "temperature", "snow", "hot", "cold"


@dataclass
class Reference(Entity):
    """Words used to refer to something, indicating possession or reference."""

    span: str  # Such as: "my", "all", "all my", "My", "my alarm"


# SKIP...

@dataclass
class MovieName(Entity):
    """The title or name of a specific movie."""

    span: str  # Such as: "The Cycle", "The Mark", "Sontha Ooru", "Health Warning", "The Belles of St. Clements"

@dataclass
class MovieType(Entity):
    """General words referring to the category of movies."""

    span: str  # Such as: "movies", "animated movies", "films", "film", "animated movie"


# This is the text to analyze
text = ""

# This is the English translation of the text
eng_text = ""

# Using translation and fusion
# (1) generate annotation for eng_text
# (2) generate annotation for text

# The annotation instances that take place in the eng_text above are listed here
result = []

# The annotation instances that take place in the text above are listed here
final_result = [
]
\end{lstlisting}
\caption{Annotation guidelines for xSID.}
\end{figure}

\begin{figure}
\begin{lstlisting}
# The following lines describe the task definition
@dataclass
class CountryRelation(Relation):
    """Country relation: Sovereign state of this item (not to be used for human beings)."""

    arg1: str
    arg2: str


@dataclass
class PlaceOfBirthRelation(Relation):
    """Place of Birth relation: most specific known (e.g., city instead of country, or
    hospital instead of city) birth location of a person, animal or fictional character.
    """

    arg1: str
    arg2: str

# SKIP...

@dataclass
class SiblingRelation(Relation):
    """Sibling relation: the subject and the object have the same parents (brother, sister, etc.)."""

    arg1: str
    arg2: str


@dataclass
class InceptionRelation(Relation):
    """Inception relation: date or point in time when the subject came into existence as defined."""

    arg1: str
    arg2: str


# This is the text to analyze
text = ""

# This is the English translation of the text
eng_text = ""

# Use both eng_text and text and generate results for text

# The list called result contains relations for the following entities:
#    - "": [Entity Type]
#
result = [
]
\end{lstlisting}
\caption{Annotation guidelines for REDFM.}
\end{figure}

\begin{figure}
\begin{lstlisting}
# The following lines describe the task definition
@dataclass
class WeatherNoun(Entity):
    """Names or terms related to weather conditions or forecasts."""

    span: str  # Such as: "weather", "temperature", "forecast", "degrees", "degree"


@dataclass
class Location(Entity):
    """Names that refer to a physical location, either a specific place name or relative location reference."""

    span: str  # Such as: "outside", "new york", "chicago", "seattle", "half moon bay"

# SKIP...

@dataclass
class NewsType(Entity):
    """Words indicating the type or category of news content."""

    span: str  # Such as: "news"


@dataclass
class WeatherTemperatureUnit(Entity):
    """Words specifying the unit used to measure temperature."""

    span: str  # Such as: "celsius", "f", "c", "celcius", "fahrenheit"


# This is the text to analyze
text = ""

# This is the English translation of the text
eng_text = ""

# Using translation and fusion
# (1) generate annotation for eng_text
# (2) generate annotation for text

# The annotation instances that take place in the eng_text above are listed here
result = []

# The annotation instances that take place in the text above are listed here
final_result = [
]
\end{lstlisting}
\caption{Annotation guidelines for MultiTO.}
\end{figure}

\begin{figure}
\begin{lstlisting}
# The following lines describe the task definition
@dataclass
class Transport(MovementEvent):
    """A Transport Event occurs whenever an Artifact (Weapon or Vehicle) or a Person is moved from one Place (GPE, Facility,
    Location) to another."""

    mention: str  # The text span that most clearly expresses (triggers) the event
    agent: List[str]  # The agent responsible for the transport Event
    artifact: List[str]  # The person doing the traveling or the artifact being traveled
    vehicle: List[str]  # The vehicle used to transport the person or artifact
    price: List[str]  # The price of transporting the person or artifact
    origin: List[str]  # Where the transporting originated
    destination: List[str]  # Where the transporting is directed
    time: List[str]  # When the transporting takes place


# This is the text to analyze
text = "..."

# This is the English translation of the text
eng_text = "..."

# Use both eng_text and text and generate results for text

# The list called result contains the instances for the following events according to the guidelines above in text:
#    - "..." triggers a [Event Type] event.
#
result = [
]

final_result = [
]
\end{lstlisting}
\caption{Annotation guidelines for ACE EAE.}
\end{figure}

\begin{figure}
\begin{lstlisting}
# The following lines describe the task definition
@dataclass
class LifeEvent(Event):
    """A LifeEvent occurs whenever a Person Entity borns, dies, gets married, divorced or gets injured."""

    mention: str
    """The text span that most clearly expresses the event.
        Such as: "wounded", "divorce", "birth", "born", "marriage" 
    """

@dataclass
class PersonnelEvent(Event):
    """A PersonnelEvent occurs when a Person entity changes its job position (JobTitle entity) with respect an
    Organization entity. It includes when a person starts working, ends working, changes offices within, gets nominated or is
    elected for a position in a Organization."""

    mention: str
    """The text span that most clearly expresses the event.
        Such as: "won", "appoint", "retired", "fired", "appointed" 
    """

# SKIP...

@dataclass
class ContactEvent(Event):
    """A ContactEvent occurs whenever two or more entities (persons or organization's representatives) come together at
    a single location and interact with one another face-to-face or directly enages in discussion via written or
    telephone communication."""

    mention: str
    """The text span that most clearly expresses the event.
        Such as: "meetings", "conference", "talked", "met", "letters" 
    """


# This is the text to analyze
text = ""

# This is the English translation of the text
eng_text = ""

# Using translation and fusion
# (1) generate annotation for eng_text
# (2) generate annotation for text

# The annotation instances that take place in the eng_text above are listed here
result = []

# The annotation instances that take place in the text above are listed here
final_result = []
\end{lstlisting}
\caption{Annotation guidelines for ACE EE.}
\end{figure}

\begin{figure}
\begin{lstlisting}
# The following lines describe the task definition
@dataclass
class PhysicalRelation(Relation):
    """The Physical Relation captures the physical location relation of entities such as: a Person entity located in a
    Facility, Location or GPE; or two entities that are near, but...
    """

    arg1: str
    arg2: str


@dataclass
class GenAffiliationRelation(Relation):
    """The GenAffiliation Relation describes the citizen, resident, religion or ethnicity relation when the `arg1` is a
    Person. When the `arg1` is an Organization, the relation ..
    """

    arg1: str
    arg2: str

# SKIP...

@dataclass
class PersonalSocialRelation(Relation):
    """The Personal-Social Relation describe the relationship between people. Both arguments must be entities of type
    Person. Please note: The arguments of these Relations are not ordered. The Relations are symmetric.
    """

    arg1: str
    arg2: str

# This is the text to analyze
text = ""

# This is the English translation of the text
eng_text = ""

# Use both eng_text and text and generate results for text

# The list called result contains the relation annotations for the following entities in text:
#    - "": [Entity Type]
#
result = []
\end{lstlisting}
\caption{Annotation guidelines for ACE RE.}
\end{figure}

\begin{figure}
\begin{lstlisting}
# The following lines describe the task definition
@dataclass
class Vehicle(Entity):
    """A Vehicle entity refers to vehicles that are used for transportation...."""

    span: str  # Such as: "ship", "car", "aircraft", "tank", "plane"

# SKIP...

@dataclass
class Facility(Entity):
    """A facility is a functional, primarily man-made structure. These include... """

    span: str  # Such as: "home", "airport", "base", "there", "hospital"

@dataclass
class GPE(Entity):
    """Geo-Political Entities are composite entities comprised of a population, a government, a physical location..."""

    span: str  # Such as: "Iraq", "U.S.", "they", "country", "Baghdad"

@dataclass
class Weapon(Entity):
    """A Weapon entity refers to instruments that can be used to deal physical damage, destroy something or kill someone. ..."""

    span: str  # Such as: "weapons", "gun", "missiles", "nuclear", "artillery"

# This is the text to analyze
text = ""

# This is the English translation of the text
eng_text = ""

# Using translation and fusion
# (1) generate annotation for eng_text
# (2) generate annotation for text

# The annotation instances that take place in the eng_text above are listed here
result = []

# The annotation instances that take place in the text above are listed here
final_result = [
]
\end{lstlisting}
\caption{Annotation guidelines for ACE NER.}
\end{figure}

\end{document}